\documentclass[]{xiaomiev}
\DeclareFontShape{T1}{xiaomiev}{b}{n}{<-> s * [1] xiaomiev/xiaomiev}{}
\DeclareFontShape{T1}{xiaomiev}{b}{it}{<-> s * [1] xiaomiev/xiaomiev}{}

\usepackage[toc,page,header]{appendix}
\usepackage{booktabs}
\usepackage{multirow}
\usepackage{graphicx}
\usepackage{array}
\usepackage{makecell}
\usepackage{amsmath}
\usepackage{amssymb}
\usepackage{placeins}
\usepackage{longtable}
\usepackage{float}
\usepackage{cuted}
\usepackage{chngcntr}
\usepackage{cleveref}
\usepackage{threeparttable}
\usepackage{tabularx}
\usepackage{subcaption}
\usepackage{wrapfig}
\usepackage{enumitem}
\usepackage{pifont}
\setlist[itemize]{leftmargin=15pt}
\renewcommand{\paragraph}[1]{\noindent\textbf{#1.}\hspace*{1em}}
\newcommand{\method}{UCAG-P}
\newcommand{\collabicon}{\raisebox{0.08ex}{\textcolor{xiaomievblue}{\scalebox{0.82}{\ensuremath{\boldsymbol{\times}}}}}}

\title{One Policy, Many Embodiments: Unified Camera-Centric Action Geometry Pre-training for Heterogeneous Embodied Manipulation}
\author{Xiaomi Embodied Intelligence Team\enspace\collabicon\enspace University of Macau}
\vspace{-2em}
\contribution{See the \protect\hyperref[sec:contributions]{\textbf{Contributions and Acknowledgments}} section for a list of contributors.}

\vspace{-2em}
\abstract{
Scaling generalist vision-language-action (VLA) policies is severely bottlenecked by the inherent heterogeneity of embodied data, which spans diverse robot morphologies, camera configurations, and low-level action spaces. Existing paradigms typically address this mismatch through explicit action retargeting, human-to-robot video synthesis, or dataset-specific adaptation branches, fundamentally hindering the joint learning of a unified policy. We introduce \method{}, a camera-centric unified action formulation that structurally aligns heterogeneous embodied datasets into a shared geometric action space. Rather than treating robot-specific commands as the shared policy target, \method{} represents manipulation through camera-observable anchor motion in camera-frame coordinates, treating robot arms, humanoids, and human hands as different embodiments of a common action schema. A geometry-conditioned action translator combines predicted motion with target-embodiment kinematics to produce executable controls. The resulting decoupled architecture allows a shared VLA policy to learn transferable manipulation geometry while retaining embodiment-specific controllability. \method{} is trained on 4.03K hours of robot and simulation data and 2.34K hours of human demonstrations. A single checkpoint reaches 98.3\% on LIBERO, 88.7\% and 89.2\% on RoboTwin Easy and Hard, 82.0\% zero-shot on LIBERO-Plus, and 62.0\% on RoboCasa GR-1, without benchmark-specific fine-tuning.
\\[2ex]
\textbf{Date:} \today\\
\textbf{Project Page \& Code:} \href{https://public-bots.github.io/UCAG-P}{\textbf{https://public-bots.github.io/UCAG-P}}\\
\vspace{-2.5em}
}

\begin{document}

\maketitle

\begin{figure}[H]
\vspace{-1em}
\centering
\includegraphics[width=0.999\textwidth]{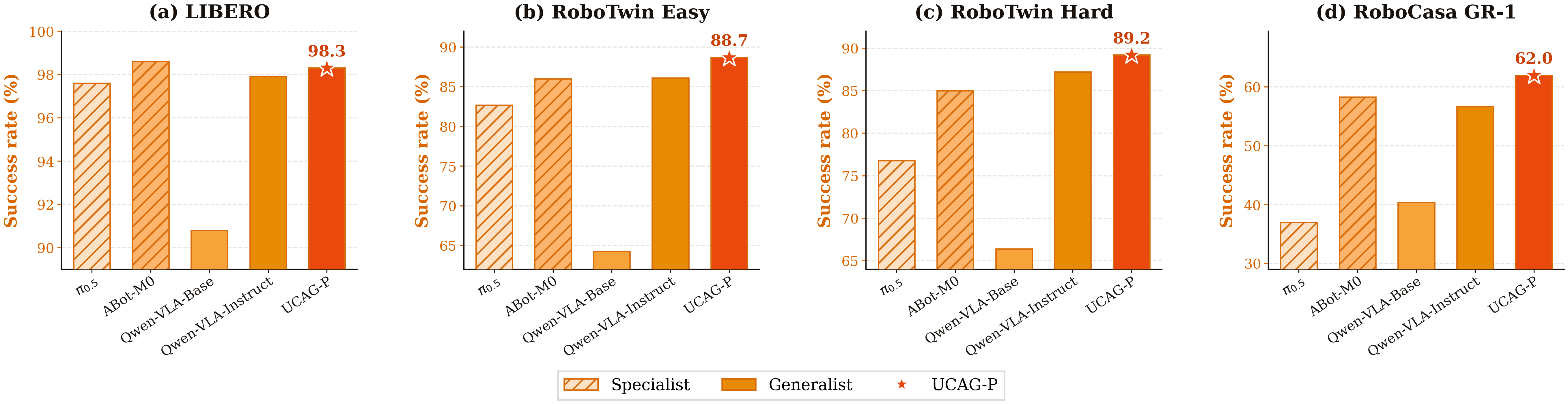}
\vspace{-2em}
\caption{\textbf{Cross-embodiment benchmark performance.} Success-rate comparison across LIBERO, RoboTwin Easy, RoboTwin Hard, and RoboCasa GR-1. \method{} uses a single unified checkpoint without benchmark-specific fine-tuning and remains competitive across single-arm, bimanual, and humanoid settings.}
\label{fig:teaser_fig}
\end{figure}

\clearpage
\onecolumn

\tableofcontents
\clearpage

\section{Introduction}

\begin{figure}[H]
\centering
\includegraphics[width=\linewidth]{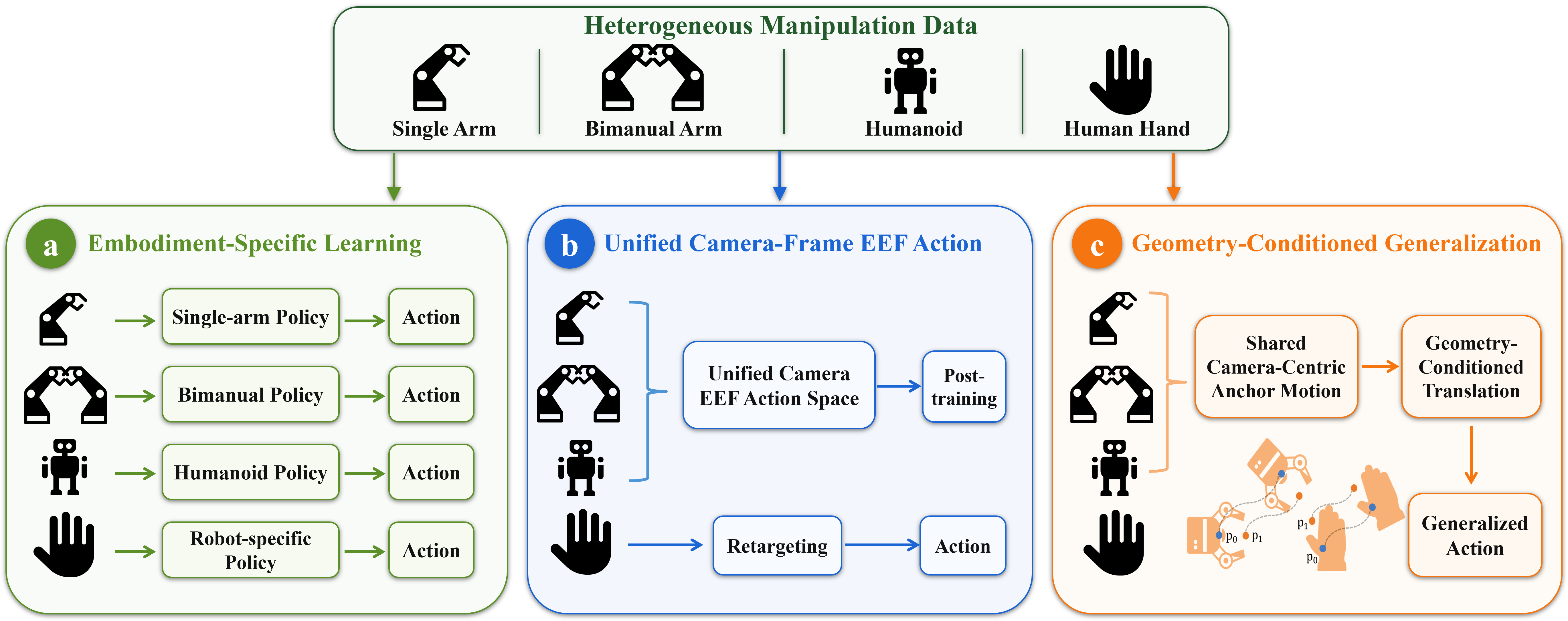}
\caption{Paradigms for heterogeneous manipulation learning. (a) Embodiment-specific policies isolate data; (b) unified camera-frame end-effector actions retain robot-specific kinematic targets; (c) \method{} shares camera-observable anchor motion before translating it into embodiment-specific commands.}
\label{fig:paradigm_overview}
\end{figure}

% \begin{wrapfigure}{r}{1\textwidth}
% \centering
% % \vspace{-4em}
% \includegraphics[width=\linewidth]{figs/intro.png}
% \caption{Paradigms for heterogeneous manipulation learning. (a) Embodiment-specific policies isolate data; (b) unified camera-frame end-effector actions retain robot-specific kinematic targets; (c) \method{} shares camera-observable anchor motion before translating it into embodiment-specific commands.}
% \label{fig:paradigm_overview}
% \end{wrapfigure}

Vision-language-action (VLA) models have emerged as a dominant paradigm for mapping language-conditioned visual observations directly to robot actions.
Recent advancements have successfully scaled these systems by leveraging larger robot datasets, large vision-language backbones, and cross-embodiment conditioning mechanisms~\citep{brohan2023rt2,kim2024openvla}.
However, the realization of generalist robot learning is bottlenecked not only by data scale but also by the inherent heterogeneity of embodied data.
Modern datasets span simulated robot trajectories, real-robot demonstrations, and human-hand videos~\citep{liu2024libero,mu2024robotwin,nasiriany2024robocasa,oneill2024openx,khazatsky2024droid}, and differ in morphology, camera geometry, proprioceptive state, control frequency, and low-level action space. As a result, the same manipulation behavior may appear as an end-effector delta, a joint command, a gripper state, or human hand motion.

Existing alignment strategies only partially address this discrepancy in action representation. As illustrated in \Cref{fig:paradigm_overview}, these strategies typically separate embodiments, retain robot-specific control targets, or rely on intermediate retargeting steps.
Benchmark-specific fine-tuning and embodiment-specific action heads avoid invalid labels, but separate action learning by dataset or robot body~\citep{kim2024openvla,black2024pi0,wang2026qwenvla}.
 Embodiment prompts and cross-embodiment conditioning provide useful context, but the predicted actions still live in embodiment-dependent control spaces \citep{oneill2024openx,wang2026qwenvla}.
Recent work such as Qwen-RobotManip improves cross-robot compatibility by aligning heterogeneous robot data through a unified camera-frame end-effector action space~\citep{qwen2026robotmanip}.
However, this interface remains centered on robot end effectors and does not directly capture human grasp behavior.
Human-to-robot retargeting, trajectory reconstruction, and video-generation pipelines can make human data usable, but they require conversion assumptions before human motion can train a robot policy~\citep{bharadhwaj2024gen2act,chen2026videomanip}.
What remains under-specified is a shared action representation that abstracts over embodiment differences while remaining grounded enough to translate into executable robot control.

\clearpage
\begin{figure}[H]
\centering
\includegraphics[width=\textwidth]{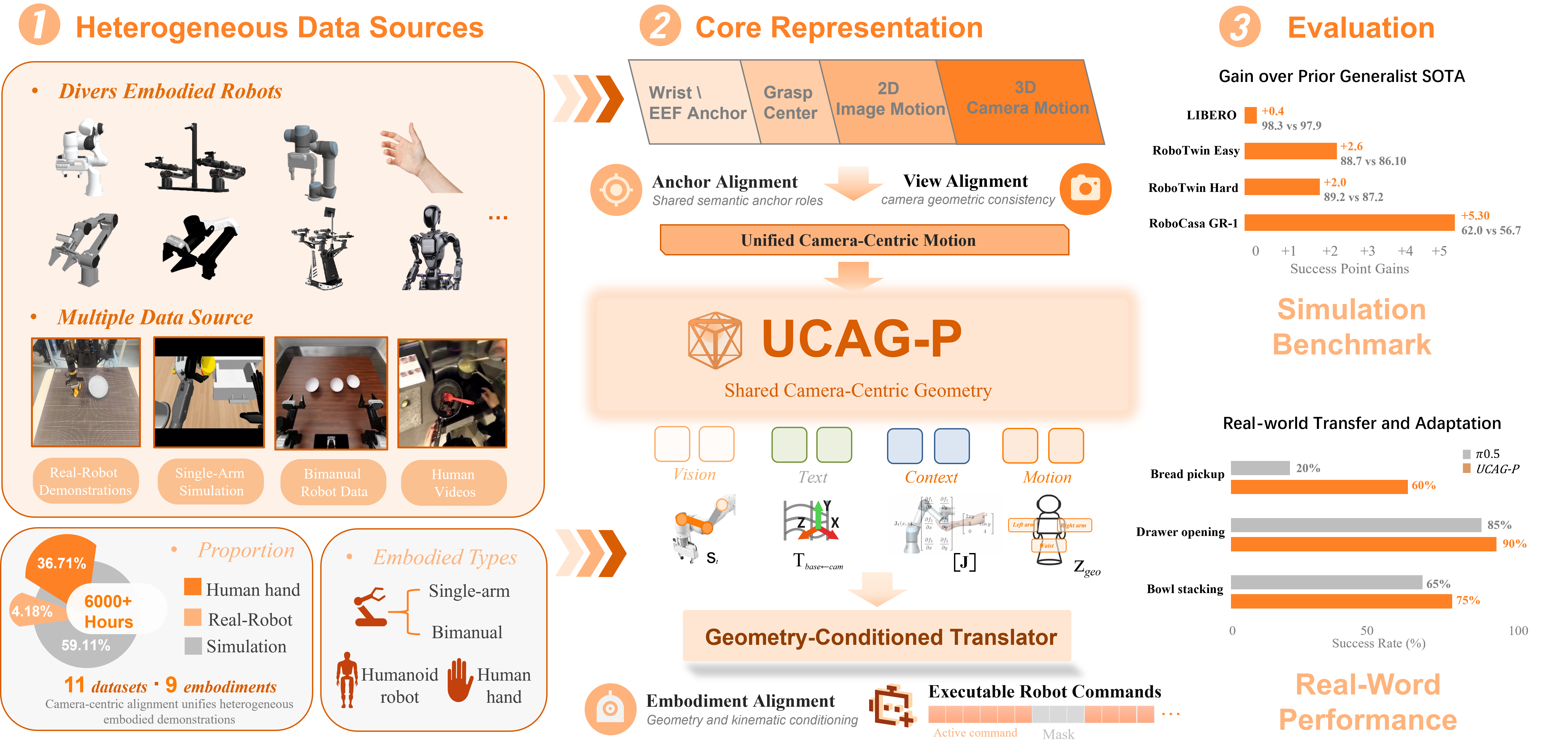}
% \vspace{-2em}
\caption{Overview of \method{} for unified learning from heterogeneous manipulation data, spanning single-arm robots, bimanual robots, dexterous robotic hands, and human hands.
\method{} maps their incompatible native action spaces into a shared camera-centric motion space defined by 3D camera-frame trajectories of the wrist or end-effector and grasp center. A geometry-conditioned translator then combines the unified motion prediction with camera-to-base transforms, Jacobians, and robot state to produce embodiment-specific executable actions.
This unified interface enables joint robot--human training and improves performance across LIBERO, LIBERO-Plus, RoboTwin, and RoboCasa GR-1.}
\label{fig:teaser}
\vspace{-1em}
\end{figure}

We argue that the most stable common structure across embodied demonstrations lies not in the low-level controller, but in the camera-observable geometry of manipulation.
Across robot arms, humanoids, and human hands, task progress is visible through the motion of the wrist or end effector and the grasp center, which can be represented in camera-frame coordinates independent of the native controller.
Building on this observation, we introduce \method{}, a camera-centric formulation for heterogeneous embodied manipulation that aligns robot and human demonstrations through shared anchor motion, as illustrated in \Cref{fig:teaser}. \method{} is trained on 11 datasets spanning more than 6,300 hours of demonstrations and nine embodiments. Concurrently, a geometry-conditioned translator maps the shared camera-centric prediction into executable commands for the target embodiment.
% \method{} treats diverse robot platforms and human demonstrations simply as different embodiments operating within a shared geometric action space. 
Such a decoupled design empowers the model to absorb multi-source data and allows massive volumes of human videos to be directly integrated into training as an independent embodiment, all while preserving precise controllability for each target platform, completely bypassing the need for computationally expensive robot-video inpainting, video editing, or explicit action retargeting as a prerequisite.

Our primary contributions are summarized as follows:
\begin{itemize}
    \item We propose a generalist manipulation policy with a novel camera-centric unified formulation for joint pre-training across heterogeneous embodied data, seamlessly integrating robot trajectories, simulation data, and human-hand demonstrations.
    % \item We treat human demonstrations as a distinct embodiment in the shared camera-centric space, allowing human-hand motion directly to supervise policy learning without robot-video synthesis or explicit retargeting.
    \item We establish a framework where human demonstrations are directly treated as a distinct embodiment within the shared action space, allowing human-hand motion directly to supervise policy learning, eliminating the need for explicit human-to-robot video synthesis or retargeting.
    \item  We introduce a decoupled, embodiment-specific action translation module that maps shared geometric actions into executable robot controls while preserving the cross-domain generalizability of the base policy.
    \item  We comprehensively validate \method{} across simulated single-arm, bimanual, humanoid, out-of-distribution, cross-embodiment, and real-robot settings, without specific dataset fine-tuning.
\end{itemize}

% \Cref{fig:teaser_fig} summarizes the cross-embodiment benchmark results. The remainder of the paper describes the pre-training corpus (\Cref{sec:pretraining-data}), method, experiments (\Cref{sec:experiments}), and results (\Cref{sec:results}), before concluding with limitations in \Cref{sec:conclusion}.

\section{Related Work}

\paragraph{Generalist VLA and Cross-Embodiment Robot Learning}
Vision-language-action (VLA) policies have become a prominent approach for language-conditioned autonomous driving and robot manipulation or navigation by connecting visual observations and language instructions to action prediction
 \citep{brohan2023rt2,kim2024openvla, qu2025eo, song2025hume,hao2025mimo,yang2026fpc,li2026spaact,li2026think,lu2026xiaomi,luo2026last,chen2026vilta,luo2026unleashing,luo2025adathinkdrive}.
Recent systems improve this paradigm through larger robot datasets, stronger multimodal backbones, open generalist policies, and more scalable action-generation mechanisms \citep{oneill2024openx,octo2024,black2024pi0,black2025fast,fang2026molmoact2,wang2026qwenvla}.
However, large-scale embodied data remain heterogeneous across morphologies, camera configurations, proprioceptive states, control frequencies, and native action spaces \citep{oneill2024openx,khazatsky2024droid,wu2024robomind,liu2024rdt}.
Prior work mitigates this mismatch with embodiment conditioning, action normalization, embodiment-specific heads, latent alignment, or canonical state-action templates \citep{kim2024openvla,kim2025openvlaoft,black2024pi0,wang2024crosslatent,yan2026unifiedlatent,wang2026qwenvla,qwen2026robotmanip}.
These approaches make important progress toward cross-embodiment robot learning, but the learned action interface often remains tied to robot-specific commands, embodiment-specific output heads, or a predefined robot-centric action schema.
\method{} follows the generalist VLA direction, but focuses on a shared action representation that can absorb heterogeneous robot and human demonstrations before translating predictions into embodiment-specific executable commands.

\paragraph{Human Demonstrations for Robot Learning}
Human videos and egocentric demonstrations provide scalable observations of object-centric manipulation, but they usually do not come with robot-native executable action labels.
Prior work has used human demonstrations for robot learning through visual pre-training, imitation learning, retargeting, trajectory reconstruction, robot-video generation, or pseudo-action extraction \citep{nair2022r3m,mandlekar2021robomimic,qin2021dexmv,bahl2022whirl,bharadhwaj2024gen2act,yu2025egomi,chen2026videomanip,shi2026egohumanoid,li2026aceego}.
These methods make human behavior usable for robot policy learning, but often require conversion assumptions about target morphology, action parameterization, or retargeting.
In contrast, 
% \method{} treats human demonstrations as another embodiment in a shared camera-centric action space.
\method{} converts human demonstrations into camera-centric motion and action supervision, and trains them together with robot trajectories in the shared action space.
This allows unpaired human and robot data to jointly supervise transferable manipulation geometry, while both converted human actions and robot actions contribute to action translation when available.

\paragraph{Action Representations for Manipulation}
The action representation determines what supervision can be shared across datasets and how policy predictions can be executed.
Existing policies use discrete tokens, continuous action chunks, flow-based or diffusion-based trajectories, end-effector deltas, waypoints, keypoints, voxel actions, and other geometric abstractions \citep{zhao2023aloha,chi2023diffusionpolicy,black2024pi0,black2025fast,zeng2020transporter,shridhar2022peract,goyal2023rvt}.
These representations are effective within their intended interfaces, but are typically tied to executable robot commands or intermediate structures for specific embodiments, task families, or observation setups.
This makes a single action target difficult to share across robots and human-hand demonstrations while still supporting executable control.
\method{} addresses this issue by separating transferable manipulation geometry from embodiment-specific execution.
It predicts camera-centric anchor motion and translates the shared prediction into the target robot's native command space, keeping the learning target common across heterogeneous data while preserving executable control.

\section{Pre-training Data}
\label{sec:pretraining-data}

The quality and diversity of training data are central to learning a policy that generalizes across embodiments, tasks, and environments. We assemble a heterogeneous pre-training corpus from three complementary sources: real-robot demonstrations, simulated robot trajectories, and egocentric human-hand videos. The embodied corpus contains 1,020,672 episodes and 6,373.586 hours across eleven dataset subsets, spanning single-arm, dual-arm, humanoid, and human-hand embodiments. A separate vision-language mixture supplements these trajectories with instruction following, spatial grounding, affordance understanding, and trajectory reasoning. \Cref{tab:data-composition} summarizes the embodied corpus, while \Cref{tab:training-data} organizes all data families by supervision and training stage. We next describe the corpus composition (\Cref{sec:embodied-corpus-composition}), data curation pipeline (\Cref{sec:data-curation}), and supervision assigned to each training stage (\Cref{sec:training-data-supervision}).

\begin{table}[H]
\centering
\setlength{\tabcolsep}{5pt}
\caption{Composition of the embodied pre-training corpus. Episode counts follow dataset metadata when available, and proportions are computed from the total number of hours.}
\label{tab:data-composition}
\begin{tabular}{llrrrl}
\toprule
Category & Dataset & Episodes & Hours & Ratio & Supervision \\
\midrule
Real robot
& RoboChallenge-HDF5~\citep{yakefu2025robochallenge} & 5,080 & 33.212 & 0.52\% & Robot action \\
& RoboCoin-Piper-LeRobot~\citep{wu2025robocoin} & 21,854 & 114.057 & 1.79\% & Robot action \\
& DROID-HDF5~\citep{khazatsky2024droid} & 26,835 & 119.079 & 1.87\% & Robot action \\
\cmidrule(lr){2-6}
& Real-robot subtotal & 53,769 & 266.348 & 4.18\% & Robot action \\
\midrule
Simulation
& RoboCasa GR-1~\citep{nasiriany2024robocasa} & 24,000 & 83.612 & 1.31\% & Robot action \\
& LIBERO-HDF5~\citep{liu2024libero} & 1,692 & 2.531 & 0.04\% & Robot action \\
& RoboTwin2-ALOHA-All~\citep{mu2024robotwin} & 2,500 & 5.114 & 0.08\% & Robot action \\
& InternData-MultiRobot~\citep{tian2025interndataa1} & 586,722 & 3649.438 & 57.26\% & Robot action \\
& InternData-Franka~\citep{tian2025interndataa1} & 10,857 & 26.816 & 0.42\% & Robot action \\
\cmidrule(lr){2-6}
& Simulation subtotal & 625,771 & 3767.511 & 59.11\% & Robot action \\
\midrule
Human hand
& VITRA-Human~\citep{vitra} & 907 & 148.727 & 2.33\% & Pseudo-action \\
& EgoDex-Human~\citep{egodex} & 338,234 & 829.000 & 13.01\% & Pseudo-action \\
& EgoVerse-Human~\citep{egoverse} & 1,991 & 1362.000 & 21.37\% & Pseudo-action \\
\cmidrule(lr){2-6}
& Human-hand subtotal & 341,132 & 2339.727 & 36.71\% & Pseudo-action \\
\midrule
Total
& All datasets & 1,020,672 & 6373.586 & 100.00\% & Mixed \\
\bottomrule
\end{tabular}
\end{table}

\begin{table*}[!ht]
\centering
\small
\setlength{\tabcolsep}{4pt}
\renewcommand{\arraystretch}{1.16}
\caption{Training data, supervision, and stage-wise use. Missing labels are masked, so each sample contributes only the targets available from its source.}
\label{tab:training-data}
\begin{tabularx}{\textwidth}{@{}
    >{\raggedright\arraybackslash}p{0.12\textwidth}
    >{\raggedright\arraybackslash}p{0.25\textwidth}
    >{\raggedright\arraybackslash}p{0.15\textwidth}
    >{\raggedright\arraybackslash}X@{}}
\toprule
\textbf{Data family} & \textbf{Representative sources} & \textbf{Embodiment} & \textbf{Supervision and training use} \\
\midrule
\textbf{Vision-language}
& ShareRobot~\citep{ji2025robobrain}; RefSpatial-v2~\citep{zhou2025roborefer}; RoboVQA~\citep{sermanet2024robovqa}; RoboAfford~\citep{tang2025roboafford}
& image--text
& \textbf{Stage 1.} Instruction following; spatial and affordance grounding; trajectory understanding; text/JSON $p_0/p_1$ coordinates. \\
\addlinespace[2pt]
\textbf{Human}
& VITRA~\citep{vitra}; EgoDex~\citep{egodex}; EgoVerse~\citep{egoverse}
& egocentric human hand
& \textbf{Stages 1--3.} RGB and language; wrist and grasp-center anchors; 3D camera-centric trajectories. \\
\addlinespace[2pt]
\textbf{Simulation}
& LIBERO~\citep{liu2024libero}; RoboTwin 2.0~\citep{mu2024robotwin}; RoboCasa GR-1~\citep{nasiriany2024robocasa}; InternData-A1~\citep{tian2025interndataa1}
& single arm, dual arm, humanoid
& \textbf{Stages 1--3.} RGB and language; robot state; calibrated $p_0/p_1$ geometry; camera-frame trajectories; executable commands. \\
\addlinespace[2pt]
\textbf{Real robot}
& DROID~\citep{khazatsky2024droid}; RoboChallenge~\citep{yakefu2025robochallenge}; RoboCoin~\citep{wu2025robocoin}
& single and dual arm
& \textbf{Stages 1--3.} RGB and language; robot state and commands; calibration when available; derived $p_0/p_1$ supervision. \\
\bottomrule
\end{tabularx}
\end{table*}

\subsection{Embodied Corpus Composition}
\label{sec:embodied-corpus-composition}

\paragraph{Robot and simulation data}
The real-robot portion combines RoboChallenge, RoboCoin, and DROID, contributing 266.348 hours of physical observations and executable commands from deployed systems. The simulated portion combines RoboCasa GR-1, LIBERO, RoboTwin 2.0, and InternData, contributing 3,767.511 hours across single-arm, dual-arm, and humanoid settings. Simulation accounts for 59.11\% of the total duration, with InternData-MultiRobot providing the largest individual share. Together, robot and simulation data constitute 63.29\% of the corpus and expose the model to varied morphology, camera geometry, scene appearance, task semantics, and control interfaces.

\paragraph{Egocentric human demonstrations}
VITRA, EgoDex, and EgoVerse contribute 2,339.727 hours of egocentric manipulation video, or 36.71\% of the embodied corpus. These videos contain no robot commands. We therefore detect wrist and grasp-center anchors and convert their motion into camera-centric pseudo-actions. This representation supplements robot trajectories with diverse object interactions and embodiment-independent motion cues while expressing human and robot behavior in the same geometric action space.

\subsection{Data Curation and Preprocessing}
\label{sec:data-curation}

The source datasets differ in storage format, camera configuration, action space, coordinate convention, and annotation granularity. We therefore convert them into a common multimodal training interface through four stages. The pipeline indexes episodes and preserves their domain identity, standardizes visual and language observations, derives camera-centric geometry and executable-control targets where available, and constructs fixed-horizon samples with validity masks. Domain identifiers and valid-dimension indicators are retained throughout, so heterogeneous embodiments can be trained jointly without treating unavailable labels or padded dimensions as supervision.

\paragraph{Stage 1: episode indexing and domain partitioning}
For each dataset, we build an episode-level index containing the sample path, domain, task instruction, available cameras, temporal stride, and mixture-sampling weight. Each episode is then parsed using its source-specific semantics. For example, LIBERO provides simulated demonstrations with absolute control targets and multiview observations; RoboCasa provides GR-1 joint states and video observations from which end-effector geometry can be recovered; RoboTwin provides bimanual end-effector poses, gripper states, and multicamera observations; and the human datasets provide egocentric video, 3D hand keypoints, and language annotations. Unreadable or incomplete episodes, trajectories that are too short, and samples missing required modalities are excluded before training.

\paragraph{Stage 2: visual and language standardization}
All RGB observations are resized to a common resolution, $224\times224$ by default, and receive lightweight color augmentation during training to reduce appearance gaps among simulation, real-robot video, and egocentric demonstrations. Because camera counts differ across sources, observations are packed into a fixed number of view slots. Missing views are zero-filled and disabled by view-validity indicators. We retain the original task instruction when available and randomly sample among annotated paraphrases to improve linguistic diversity.

\paragraph{Stage 3: geometric and action alignment}
Robot supervision is derived from the most reliable state and control signals available in each source. End-effector or wrist poses, gripper geometry, camera calibration, and forward kinematics are used to recover the semantic anchors $p_0$ and $p_1$ in camera coordinates. LIBERO targets are aligned with scene-specific camera parameters; RoboTwin bimanual trajectories are transformed into the training reference frame; and RoboCasa joint states are decomposed by body part and passed through cached forward kinematics or the simulator to recover both arm anchors and camera geometry. Single-arm and bimanual samples are packed into a common layout, with inactive arms and unavailable coordinates masked. For human demonstrations, the wrist keypoint defines $p_0$ and the midpoint of the thumb and index fingertips defines $p_1$; robot-only quantities such as executable commands are left invalid rather than synthesized.

\paragraph{Stage 4: temporal windowing and normalization}
Each processed trajectory is converted into short-horizon samples anchored at the current frame. Given a trajectory $\tau$ and horizon $H$, the current observation is taken from $\tau_t$ and the following steps form the supervision chunk $\{g_{t+h}\}_{h=1}^{H}$. We use $H=30$ by default, while temporal stride and sampling frequency remain domain specific. Short terminal segments are padded to the target length and excluded from the loss by temporal-validity masks. When enabled, end-effector and joint quantities are scaled with domain-specific statistics; otherwise, they remain in processed physical units. This procedure yields a consistent batch structure while preserving each source's valid geometric and control targets.

\begin{figure}[H]
\centering
\includegraphics[width=\textwidth]{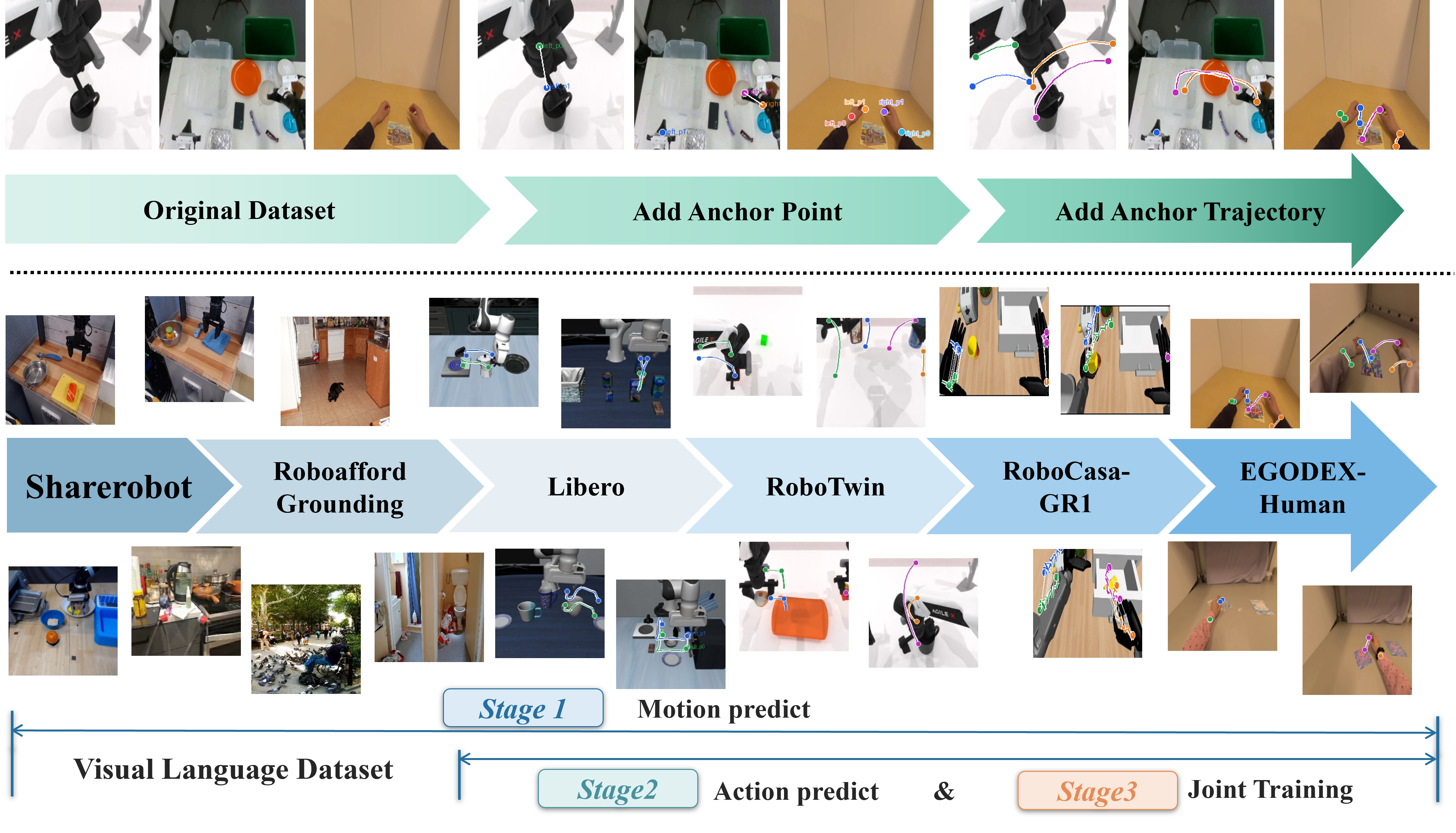}
\vspace{-1.5em}
\caption{\textbf{Data organization and three-stage training pipeline}. Vision-language and human demonstrations establish shared visual-geometric prediction, while simulation and real-robot data additionally supervise geometry-conditioned action translation and executable control.}
\label{fig:data-pipeline}
\end{figure}

\subsection{Supervision and Training Roles}
\label{sec:training-data-supervision}

The data families provide complementary supervision. Vision-language samples provide instruction following, spatial and affordance grounding, trajectory understanding, and text/JSON coordinates during Stage 1. Because these image--text sources are not organized as temporally grounded embodied trajectories, measuring their scale in hours is not meaningful; accordingly, the hour-based statistics in \Cref{tab:data-composition} include only embodied datasets. Human data provide RGB observations, language, anchor locations, and camera-centric pseudo-trajectories. Simulation and real-robot data additionally provide robot states, calibration, and executable commands.
As illustrated in \Cref{fig:data-pipeline}, these supervision types support the progressive training design: Stage 1 establishes shared visual-geometric prediction, Stage 2 learns geometry-conditioned action translation, and Stage 3 jointly optimizes camera-centric motion and robot control across heterogeneous embodiments. Each sample contributes only the targets available from its source, with missing labels excluded through loss masks.

\section{Method}

\subsection{Problem Formulation}

We consider a mixture of embodied manipulation datasets $\mathcal{D}=\{\mathcal{D}^{(k)}\}_{k=1}^{K}$, where each dataset $\mathcal{D}^{(k)}$ is associated with its embodiment, sensor configuration, and control interface.
A trajectory is denoted as:
\begin{equation}
\tau=\{(o_t, l, s_t^{(k)}, a_t^{(k)})\}_{t=1}^{T}
\end{equation} 
where $o_t$ is the visual observation, $l$ is the language instruction, $s_t^{(k)}$ is an optional embodiment-specific state, and $a_t^{(k)} \in \mathcal{A}^{(k)}$ is the native action label. 
The native action spaces $\mathcal{A}^{(k)}$ are generally incompatible.
When the native label is executable by embodiment $k$, we denote the corresponding command chunk by $u_{t:t+H}^{(k)}\in(\mathcal{U}^{(k)})^{H}$.
Across datasets, the same manipulation intent may be annotated as end-effector deltas, joint-position commands, humanoid commands, or human wrist-hand motion.
A single low-level command parameterization therefore cannot serve as the shared prediction target for all data sources.

Our objective is to \textit{\textbf{pre-train a general policy}} that can learn from heterogeneous action annotations \textit{\textbf{without requiring every dataset to share the same executable control space}}.
\method{} separates this objective into shared geometric prediction and embodiment-specific execution.
It learns a shared camera-centric action space $\mathcal{G}$ and a base policy $\pi_\theta(o_t,l)=\hat{g}_{t:t+H}$, where $\hat{g}_{t:t+H}\in\mathcal{G}^{H}$ is a geometric action chunk over horizon $H$.
For a target embodiment that requires low-level execution, our framework translates the shared prediction into an executable command chunk $\hat{u}_{t:t+H}^{(k)}\in(\mathcal{U}^{(k)})^{H}$ in the embodiment's native control space.
In this decomposition, camera-observable manipulation motion serves as the shared policy target. A geometry-conditioned translation stage then generates embodiment-specific commands for execution.

\begin{figure}[t]
\centering
\includegraphics[width=\linewidth]{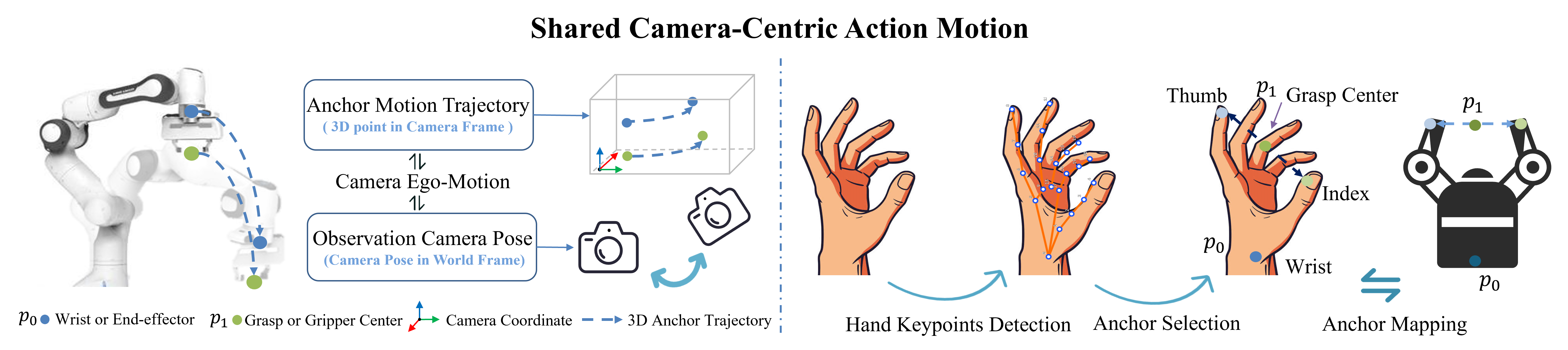}
\vspace{-1em}
\caption{Overview of shared camera-centric action space and definition of semantic anchor pair.}
\label{fig:shared_space}

\end{figure}

\subsection{Camera-Centric Action Space}

The shared space $\mathcal{G}$ describes manipulation through camera-observable geometry rather than embodiment-specific controls, as presented in ~\Cref{fig:shared_space}.
Across robot arms, humanoid manipulators, and human demonstrations, task progress is visible through the motion of the acting wrist or end effector and the grasp center in the camera view.
These quantities are tied to the visual evidence used by a VLA policy, but they are not tied to a particular joint layout or controller.
We represent manipulation with a semantic anchor pair $(p_0,p_1)$.
The anchor $p_0$ denotes the wrist or end-effector point, and $p_1$ denotes the gripper center for robot data or the grasp center for human-hand data.
For each anchor $p_i$, we use its camera-frame coordinate $p_{i,t}^{\mathrm{cam}}\in\mathbb{R}^{3}$.
At time $t$, the ground-truth camera-centric action chunk is denoted as $g_{t:t+H}=\{g_{t+h}\}_{h=1}^{H}$ over horizon $H$, and the base policy predicts $\hat{g}_{t:t+H}$.
Each step describes anchor motion in 3D camera-frame coordinates, optionally together with camera motion:
\begin{equation}
     {g}_{t+h}
    =
    \left[
    \mathbf{m}^{L}_{t+h},
    \mathbf{m}^{R}_{t+h},
    \xi_{t+h}^{\mathrm{cam}}
    \right]
    \in \mathbb{R}^{30},
\end{equation}
For each manipulator $a\in\{L,R\}$, we define
$  \mathbf{m}^{a}_{t+h} =
    \left[
        \Delta p_{0,t+h}^{\mathrm{cam}},
        \Delta p_{1,t+h}^{\mathrm{cam}},
        \Delta\psi_{t+h}^{\mathrm{cam}},
        \gamma_{t+h}^{\mathrm{gripper}},
    \mathbf{0}
    \right]
    \in\mathbb{R}^{10}
$, where $\Delta\psi_{t+h}^{\mathrm{cam}}$ is in-plane rotation, $\Delta \gamma_{t+h}^{\mathrm{gripper}}$ is gripper state or hand openness, and $\mathbf{0}$ is a padding vector to fill the 10-dimensional slot.
All displacement terms are measured relative to the corresponding anchor coordinate at \textit{\textbf{the first frame of the chunk}}.
The camera-motion term $\xi_{t+h}^{\mathrm{cam}}$ is used only when the observing camera moves and is set to zero for fixed-camera datasets, while the camera-frame displacements preserve metric motion, depth, and scale.
For multi-manipulator samples, the same anchor layout is repeated for each active manipulator, and unavailable anchors or coordinates are excluded by masks, more details can be found in \Cref{app:sparse-command-layout}.

The two-anchor representation keeps the semantic role of each anchor fixed while allowing its physical instantiation to vary across embodiments.
For robot embodiments, $p_0$ is obtained from the wrist or end-effector frame provided by simulator state or forward kinematics, and $p_1$ is computed from the gripper or hand geometry.
For human-hand demonstrations, hand keypoints provide the corresponding anchor pair, with $p_0$ obtained from the wrist keypoint and $p_1$ defined as the midpoint between the thumb tip and index fingertip.
When metric depth or camera calibration information is unavailable, the corresponding camera-frame components are marked unavailable rather than treated as valid geometric targets. By decoupling visual geometry from native controllers, this approach allows heterogeneous data to optimize a unified action space, with missing labels bypassed via loss masking.
% Robot trajectories provide anchor motion through state or kinematics, while human demonstrations provide analogous hand-anchor motion.
% Heterogeneous demonstrations can therefore supervise the same camera-centric action space whenever the corresponding geometric labels are available, with missing components handled by masks during training.

% \input{sections/30d_motion_representation.tex}

\begin{figure*}[ht]
\centering
\includegraphics[width=\textwidth]{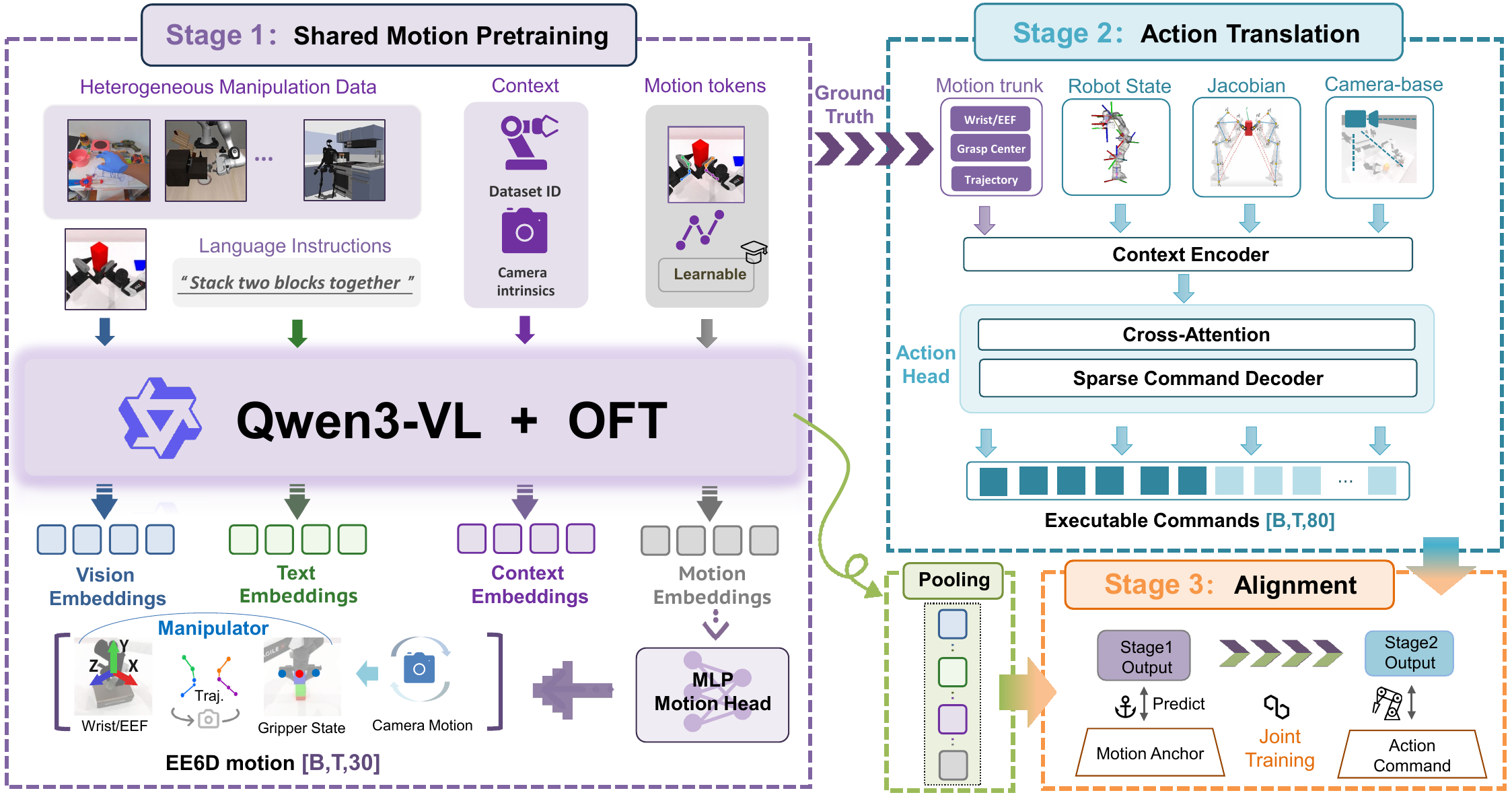}
\caption{Overview of \method{}. We train a unified VLA policy on heterogeneous embodied data spanning single-arm robots, bimanual robots, humanoids, and human-hand datasets. \method{} converts these data sources into a shared camera-centric $p_0/p_1$ action schema and learns to predict anchor motion in camera-frame coordinates. A geometry-conditioned translator maps the shared motion prediction into executable commands for the target embodiment, allowing one policy to support heterogeneous robot control and human-to-robot transfer without explicit action retargeting or robot-video synthesis.}
\label{fig:architecture}

\end{figure*}

\subsection{\method{} Architecture}

\method{} comprises a vision-language backbone for encoding observations and language, a shared camera-centric motion head to predict camera-centric motion, and a geometry-conditioned action translator that converts the predicted motion into executable commands for the target embodiment, as illustrated in \Cref{fig:architecture}.

\paragraph{Shared camera-centric policy}
We instantiate the backbone with Qwen3-VL-4B-Instruct~\citep{bai2025qwen3vl}.
Multi-view RGB observations, the language instruction, and optional proprioceptive context are encoded into a sequence of hidden states. Additionally, we append a set of learnable action-query tokens to the input sequence, where the final-layer hidden states serve as compact features for action prediction.
The camera-centric motion head maps the action-query features to the shared geometric action chunk $\hat{g}_{t:t+H}$.
It is shared across all data sources and is supervised whenever the corresponding camera-centric labels are available.

\paragraph{Geometry-conditioned translation}
The camera-centric chunk specifies motion in the observing camera frame, but target robots execute commands in embodiment-specific control spaces.
The translator conditions on robot-camera geometry and local kinematics to map the shared proposal into the command interface required by the target robot.

For embodiment $k$, the translator receives the predicted camera-centric chunk $\hat{g}_{t:t+H}$, the current embodiment state $s_t^{(k)}$, and a geometric context.
This context contains the camera-to-base transform $T_{\mathrm{base}\leftarrow\mathrm{cam}}^{(k)}$, local Jacobians $J_t^{(k)}$ evaluated at the current state, and geometry tokens $z_{\mathrm{geo}}^{(k)}$ that encode embodiment-level structure, such as active manipulators and command slots:
\begin{equation}
\hat{u}_{t:t+H}^{(k)} =  \phi_\psi\left(\hat{g}_{t:t+H}, s_t^{(k)},
        T_{\mathrm{base}\leftarrow\mathrm{cam}}^{(k)}, J_t^{(k)}, z_{\mathrm{geo}}^{(k)} \right).
\end{equation}

The camera-to-base transform aligns directions and displacements predicted in the camera frame with the robot body frame.
The Jacobian provides a local relation between joint changes and end-effector motion, which helps the translator realize a geometric displacement under the current configuration.
The geometry tokens allow the translator to condition its output on different robot bodies.
The translator outputs an executable command chunk $\hat{u}_{t:t+H}^{(k)}$ in the sparse command layout.
Only the command slots associated with the target embodiment are passed to the downstream controller.

\paragraph{Sparse executable command layout}
For embodiments that require executable joint commands, \method{} outputs a command chunk in a unified sparse command layout. In our implementation, this layout uses an 80-dimensional vector whose active slots depend on the target embodiment. Inactive dimensions are masked from the loss and are not sent to the downstream controller.
The slot assignment used in our experiments is listed in \Cref{app:sparse-command-layout}.

\subsection{Three-Stage Training}
\label{sec:three-stage-training}

We train \method{} in three stages, as shown in \Cref{fig:architecture}.
The first two stages learn shared geometric prediction and embodiment-specific translation separately, and the final stage optimizes them jointly.

\paragraph{Stage 1: Camera-centric specialization}
We first train the VLM backbone and shared camera-centric motion head using all samples with available camera-centric supervision. This stage establishes anchor-motion prediction in camera-frame coordinates as the common action interface across robot and human demonstrations. The architecture is detailed in \Cref{app:head-architectures}.

\paragraph{Stage 2: Geometry-conditioned action translation}
We then train the geometry-conditioned translator using ground-truth camera-frame trajectories, target-embodiment state, camera-to-base calibration, local Jacobians, and executable command labels.
Using ground-truth trajectories isolates the geometry-to-command mapping from upstream motion-prediction errors.
Only active command dimensions for the current embodiment contribute to the translator loss. The architecture is detailed in \Cref{app:head-architectures}.

\paragraph{Stage 3: Joint robot-human training}
Finally, we jointly optimize the shared motion head and the translator on the mixed robot-human corpus.
The translator receives trajectories predicted by the shared policy, which matches its training input to the inference-time input.
To efficiently condition the translator on VLM context, we use learnable query tokens to attention-pool the full hidden-state sequence from the last layer and pass the pooled features to the action head to preserve global visual-language context. Furthermore, auxiliary general VLM data are excluded from this stage so that the final checkpoint focuses on embodied control.

\subsection{Training objectives and Inference}
We trained our \method{} with masked objectives, allowing heterogeneous samples to supervise only the camera-centric and executable-command targets they provide.
For the shared camera-centric branch, we use a masked L1 loss over the geometric action chunk:
\begin{equation}
\mathcal{L}_{geo} = \frac{
\sum_{h=1}^{H}\sum_{d} m_{h,d}^{g}
\left|\hat{g}_{t+h,d}-g_{t+h,d}\right|}{
\max(1,\sum_{h=1}^{H}\sum_{d}m_{h,d}^{g})},
\end{equation}
where $m_{h,d}^{g}$ indicates whether dimension $d$ of the geometric target is available.
When executable command labels are available, the translator is trained with a masked L1 loss over the sparse command layout:
\begin{equation}
\mathcal{L}_{cmd} = \frac{
\sum_{h=1}^{H}\sum_{d} m_{h,d}^{u}
\left|\hat{u}_{t+h,d}^{(k)}-u_{t+h,d}^{(k)}\right|}{
\max(1,\sum_{h=1}^{H}\sum_{d}m_{h,d}^{u})},
\end{equation}
where $m_{h,d}^{u}$ masks command dimensions that are inactive for the current embodiment.
The total objective is:
\begin{equation}
\mathcal{L} = \mathcal{L}_{geo} + \lambda_{cmd}\mathcal{L}_{cmd},
\end{equation}
where $\lambda_{cmd}$ balances camera-centric geometric supervision and executable command supervision.

At inference time, the base policy first predicts $\hat{g}_{t:t+H}$ from the current observation and instruction.
For executable robot control, the translator conditions on the target embodiment state and geometry to produce $\hat{u}_{t:t+H}^{(k)}$.
The downstream controller executes only the active command slots for the current embodiment.

\section{Experiments}
\label{sec:experiments}

We evaluate whether \textbf{a single unified \method{} checkpoint} can (i) operate across heterogeneous embodiments and control interfaces, (ii) remain robust under distribution shift, and (iii) transfer to new embodiments and physical robots.
Our evaluation spans single-arm, bimanual, humanoid, out-of-distribution, cross-embodiment, and real-world settings. 
% \Cref{sec:experimental-setup} describes the training data and implementation, while \Cref{sec:results} presents the simulation and real-world results.

\subsection{Experimental Setup}
\label{sec:experimental-setup}

\paragraph{Training data}
We pre-train \method{} on the heterogeneous corpus described in \Cref{sec:pretraining-data}, which combines real-robot demonstrations, simulated trajectories, egocentric human-hand videos, and vision-language supervision.
\paragraph{Implementation details}
We instantiate \method{} with the Qwen3-VL-4B-Instruct backbone~\citep{bai2025qwen3vl}.
Unless otherwise stated, every simulation benchmark uses the same final checkpoint from the three-stage training procedure in \Cref{sec:three-stage-training}, without benchmark-specific fine-tuning.
The optimization schedule, benchmark settings, baselines, and evaluation protocols are reported in \Cref{appendix:training-evaluation-details}.

\subsection{Experimental Results}
\label{sec:results}
We evaluate \method{} across heterogeneous simulation datasets and real-world manipulation tasks. \textit{All results presented in this section are achieved using a single unified checkpoint.} 
% After evaluating unified manipulation and robustness, we examine transfer to new embodiments and real-world adaptation in \Cref{sec:generalization-transfer-results}.

\begin{table*}[ht]
\centering
\caption{Unified cross-embodiment manipulation success rates (\%). "Specialist" denotes methods post-trained on a specific benchmark. 
Bold and underline indicate the best and second-best results in each column, respectively.
% Bold: best; underline: second best
% Parenthesized values report absolute percentage-point gains over Qwen-VLA-Instruct.
}
\vspace{-1em}
\label{tab:main-results}
% \small
% \setlength{\tabcolsep}{5pt}
% 
\begin{tabular}{lccccc}
\toprule
\multicolumn{2}{c}{\scriptsize\itshape \underline{Full results are provided in \Cref{app:extended-results}.}} & \llap{$\lhd$\hspace{10pt}}Single-arm & \multicolumn{2}{c}{Bimanual} & Humanoid \\
\cmidrule(lr){3-3}\cmidrule(lr){4-5}\cmidrule(lr){6-6}
Method & Type & LIBERO & \shortstack{RoboTwin\\Easy} & \shortstack{RoboTwin\\Hard} & \shortstack{RoboCasa\\GR-1} \\
\midrule
$\pi_0$~{\scriptsize\citep{black2024pi0}} & Specialist & 94.4 & 65.9 & 58.4 & -- \\
StarVLA-OFT~{\scriptsize\citep{starvla2026}} & Specialist & 96.6 & 50.4 & -- & 48.8 \\
GR00T N1.6~{\scriptsize\citep{bjorck2025groot}} & Specialist & 97.2 & 47.6 & -- & 49.9 \\
$\pi_{0.5}$~{\scriptsize\citep{physicalintelligence2025pi05}} & Specialist & 97.6 & 82.7 & 76.8 & 37.0 \\
ABot-M0~{\scriptsize\citep{yang2026abotm0}} & Specialist & \underline{98.6} & 86.0 & 85.0 & 58.3 \\
Being-H0.5~{\scriptsize\citep{luo2026beingh05}} & Specialist & 97.6 & -- & -- & 53.3 \\
Being-H0.7~{\scriptsize\citep{luo2026beingh07}} & Specialist & \textbf{99.2} & \textbf{90.2} & \textbf{89.6} & -- \\
OASIS~{\scriptsize\citep{chen2026oasis}} & Specialist & 97.6 & -- & -- & -- \\
MolmoAct2-Think~{\scriptsize\citep{fang2026molmoact2}} & Specialist & 98.1 & -- & -- & -- \\
DeMaVLA~{\scriptsize\citep{su2026demavla}} & Specialist & -- & 88.4 & 86.8 & -- \\
LingBot-VLA~{\scriptsize\citep{wu2026lingbotvla}} & Specialist & -- & 88.5 & 86.7 & -- \\
JoyAI-RA~{\scriptsize\citep{zhang2026joyaira}} & Specialist & -- & -- & -- & \underline{63.2} \\
ZR-0~{\scriptsize\citep{li2026zr0}} & Specialist & 97.8 & \underline{88.7} & 87.9 & \textbf{69.3} \\
\midrule
Qwen-VLA-Base~{\scriptsize\citep{wang2026qwenvla}} & Generalist & 90.8 & 64.3 & 66.4 & 40.4 \\
Qwen-VLA-Instruct~{\scriptsize\citep{wang2026qwenvla}} & Generalist & 97.9 & 86.1 & 87.2 & 56.7 \\
\method{} & Generalist & 98.3~{(+0.4)} & \underline{88.7}~{(+2.6)} & \underline{89.2}~{(+2.0)} & 62.0~{(+5.3)} \\
\bottomrule
\end{tabular}

\end{table*}

\paragraph{Unified Manipulation Across Embodiments}
The first question is whether one policy can remain effective when the embodiment and command interface change.
\Cref{tab:main-results} shows that \method{} preserves strong single-arm performance while remaining competitive in the more heterogeneous bimanual and humanoid settings.
On LIBERO, \method{} reaches 98.3\%, within one point of the strongest reported result, with 98.8\%, 98.6\%, 99.2\%, and 96.4\% on LIBERO-Spatial, LIBERO-Object, LIBERO-Goal, and LIBERO-Long, respectively (see \Cref{sec:libero-suite-appendix}).
On RoboTwin, it reaches 88.66\% on Easy and 89.20\% on Hard, remaining competitive with recent specialist and generalist policies.
%  and obtaining the second-highest Hard result in the table.
On RoboCasa GR-1, where the policy must produce executable humanoid commands, \method{} reaches 62.0\%, below ZR-0's 69.3\% and JoyAI-RA's 63.2\% but above the earlier baselines included in the table~\citep{li2026zr0,zhang2026joyaira}.
\method{} outperforms the generalist baseline, Qwen-VLA, across all evaluated datasets. {While specialist models fine-tuned on single benchmarks achieve slightly higher peak metrics, our framework operates strictly without any dataset-specific post-training or fine-tuning.}
% These results support the central design goal of \method{}: the shared policy predicts camera-centric manipulation geometry, while the translator adapts this prediction to the active embodiment. 

% \begin{figure*}[t]
% \centering
% \includegraphics[width=\textwidth]{figs/libero_plus_results.pdf}
% \caption{Out-of-distribution robustness across all seven LIBERO-Plus perturbation categories. Total summarizes Camera, Robot, Language, Light, Background, Noise, and Layout success rates.}
% \label{fig:libero-plus-results}
% \end{figure*}

\begin{table*}[t]
\centering
\caption{LIBERO-Plus success rates (\%) across seven perturbation categories. Baselines results are from ABot-M0~\citep{yang2026abotm0}, "*" denote the methods are post-trained on single standard LIBERO dataset but without fine-tuning on LIBERO-Plus.}
\label{tab:libero-plus-breakdown}
% \small
% \setlength{\tabcolsep}{5pt}
% . \method{} is evaluated zero-shot, which pre-trained on heterogeneous data without specific dataset post-training
\begin{tabular}{llcccccccc}
\toprule
Method & Setting & Camera & Robot & Lang. & Light & Bkg. & Noise & Layout & Avg. \\ \hline
OpenVLA~{\scriptsize\citep{kim2024openvla}} & Zero-shot* & 0.8 & 3.5 & 23.0 & 8.1 & 34.8 & 15.2 & 28.5 & 15.6 \\
OpenVLA-OFT & Zero-shot* & 56.4 & 31.9 & 79.5 & 88.7 & 93.3 & 75.8 & 74.2 & 69.6 \\
OpenVLA-OFT-w & Zero-shot* & 10.4 & 38.7 & 70.5 & 76.8 & 93.6 & 49.9 & 69.9 & 55.8 \\
OpenVLA-OFT-m & Zero-shot* & 55.6 & 21.7 & 81.0 & 92.7 & 91.0 & 78.6 & 68.7 & 67.9 \\
NORA~{\scriptsize\citep{hung2025nora}} & Zero-shot* & 2.2 & 37.0 & 65.1 & 45.7 & 58.6 & 12.8 & 62.1 & 39.0 \\
WorldVLA~{\scriptsize\citep{cen2025worldvla}} & Zero-shot* & 0.1 & 27.9 & 41.6 & 43.7 & 17.1 & 10.9 & 38.0 & 25.0 \\
UniVLA~{\scriptsize\citep{bu2025univla}} & Zero-shot* & 1.8 & 46.2 & 69.6 & 69.0 & 81.0 & 21.2 & 31.9 & 42.9 \\
$\pi_0$~{\scriptsize\citep{black2024pi0}} & Zero-shot* & 13.8 & 6.0 & 58.8 & 85.0 & 81.4 & 79.0 & 68.9 & 53.6 \\
$\pi_0$\mbox{-Fast}~{\scriptsize\citep{black2025fast}} & Zero-shot* & 65.1 & 21.6 & 61.0 & 73.2 & 73.2 & 74.4 & 68.8 & 61.6 \\
RIPT-VLA~{\scriptsize\citep{tan2025interactive}} & Zero-shot* & 55.2 & 31.2 & 77.6 & 88.4 & 91.6 & 73.5 & 74.2 & 68.4 \\
ABot-M0~{\scriptsize\citep{yang2026abotm0}} & Zero-shot* & 60.4 & 67.9 & 86.4 & 96.2 & 91.6 & 86.4 & 82.6 & 80.5 \\
\midrule
\textbf{\method{}} & \textbf{Zero-shot} & \textbf{51.2} & \textbf{92.8} & \textbf{83.5} & \textbf{98.9} & \textbf{98.0} & \textbf{75.2} & \textbf{74.3} & \textbf{82.0} \\
\bottomrule
\end{tabular}

\end{table*}

\paragraph{Robustness Under Distribution Shift}
The second question is whether the shared geometric interface remains stable when the distribution changes.
% LIBERO-Plus perturbs the standard LIBERO setting along camera, robot-state, language, lighting, background, noise, and layout dimensions.
\Cref{tab:libero-plus-breakdown} compares \method{} with methods for which all seven LIBERO-Plus perturbation results are available. \method{} obtains a zero-shot total score of 82.0\%.
% Under the zero-shot LIBERO-Plus protocol reported by ABot-M0, \method{} is above the listed baselines, including $\pi_0$ and ABot-M0.
% As shown in \Cref{fig:libero-plus-results}, 
which is particularly robust to robot-state, lighting, and background perturbations, while camera, sensor-noise, and object-layout shifts remain more challenging. \textbf{Notably, our framework is entirely zero-shot and requires no LIBERO dataset for post-training.}

\begin{figure}[h]
\centering
\includegraphics[width=0.8\columnwidth]{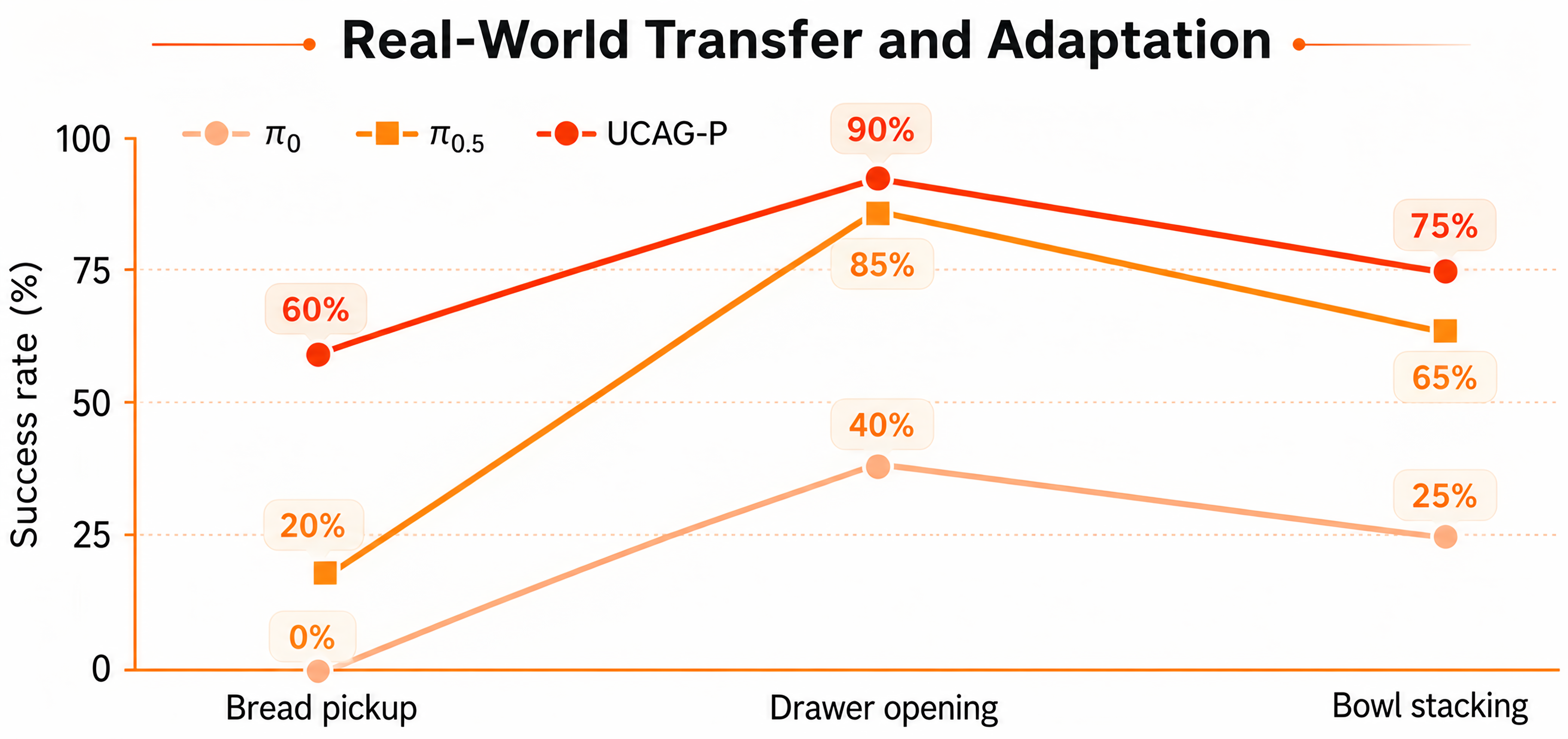}
\caption{Real-world success rates on Piper robots. Bread pickup evaluates human-to-robot transfer, while drawer opening and bowl stacking use robot demonstrations; bowl stacking additionally requires bimanual hand detection.}
\label{fig:real-world-results}
\end{figure}

\paragraph{New Embodiment Transfer and Adaptation}
\label{sec:generalization-transfer-results}
\indent\textbf{a) Real-World} The real-world study asks whether the same formulation remains useful outside simulation. 
% Bread pickup evaluates human-to-robot transfer from hand demonstrations, while drawer opening and bowl stacking evaluate SFT adaptation from robot demonstrations.
As shown in \Cref{fig:real-world-results},
% On bread pickup, both $\pi_0$ and $\pi_{0.5}$ obtain 0\%, while \method{} reaches 75\%.
% This result indicates that MediaPipe-derived camera-centric hand anchors can provide useful supervision for robot execution.
\method{} achieves 60\%, 90\%, and 75\% success on bread grasping, drawer opening, and bowl stacking, respectively, compared with 20\%, 85\%, and 65\% for $\pi_{0.5}$.
These controlled comparisons show that the \method{} initialization improves real-robot adaptation under the same demonstration budget, observation setup, and controller.
\textbf{b) Simulation} We next isolate embodiment shift by replacing directly the ALOHA robot in RoboTwin with ARX. % The scenes, object layouts, perturbation seeds, and camera extrinsics are fixed, and the initial end-effector poses are aligned by inverse kinematics.
% Under this protocol, the main change is the robot morphology and kinematic structure.
\method{} achieves 35.0\% success, which is substantially below the source-embodiment performance, but it is still zero-shot transfer under a direct embodiment replacement. The result suggests that camera-centric motion provides a transferable intermediate target, while also showing that morphology and kinematic mismatch remain open challenges. More details are provided in \Cref{sec:generalization-transfer}.

\begin{figure*}[h]
\centering
\includegraphics[width=\textwidth]{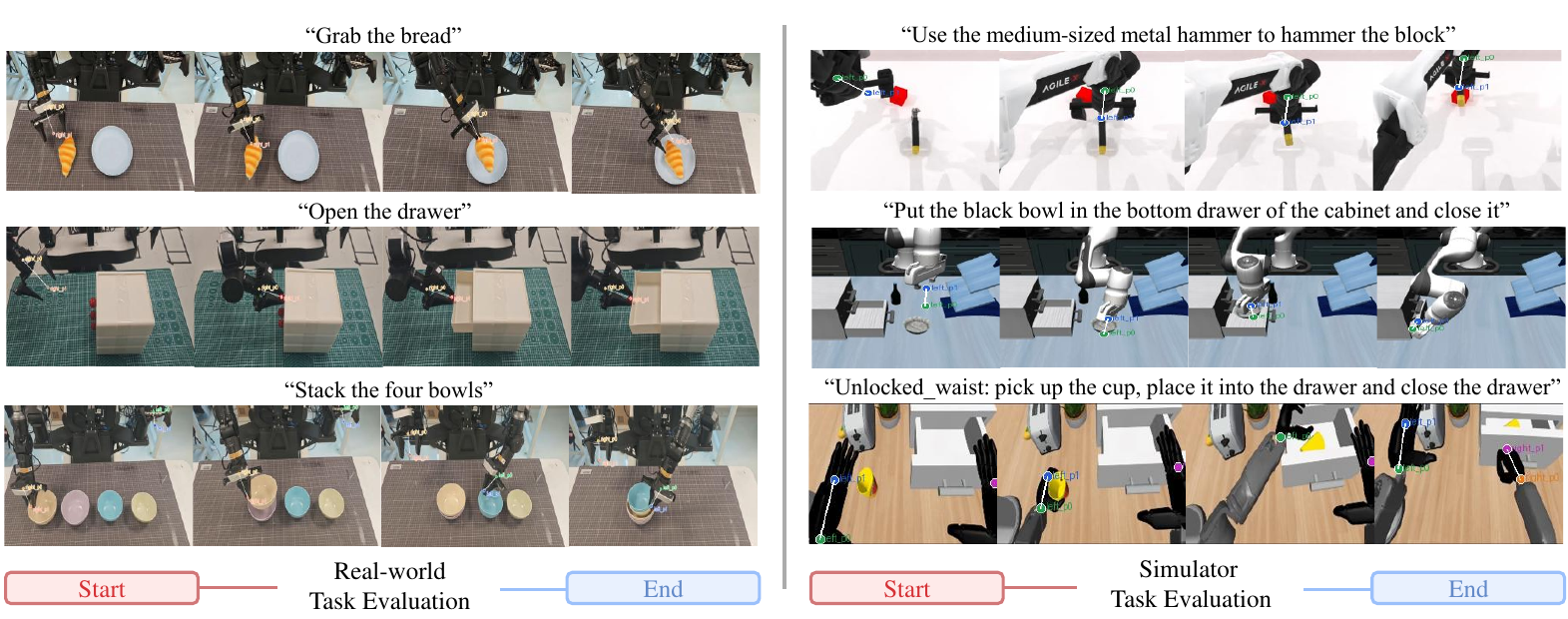}
\caption{Representative start-to-end task executions in the real world (left) and simulation (right). The real-world tasks include bread grasping, drawer opening, and bimanual bowl stacking; the simulation tasks include hammering, bowl placement with drawer closing, and waist-unlocked humanoid cup placement.}
\label{fig:task}
\end{figure*}

\paragraph{Task Visualizations}
\Cref{fig:task} qualitatively demonstrates that the shared camera-centric representation preserves task-relevant geometry across real-world and simulation domains. The successful transitions from initial observations to completed states indicate that the policy can maintain coherent object--effector relationships despite changes in scene appearance, robot morphology, and control interface. In particular, the bimanual and humanoid cases suggest that the geometry-conditioned translator can convert shared motion predictions into coordinated embodiment-specific commands. Additional qualitative examples are provided in \Cref{app:qualitative-examples}.

\section{Conclusion and Limitations}
\label{sec:conclusion}
We presented \method{}, a unified camera-centric formulation for learning vision-language-action pre-training policies from heterogeneous embodied data.
By representing manipulation through camera-observable anchor motion and translating the shared geometric prediction into embodiment-specific commands, \method{} decouples transferable task geometry from robot-specific control interfaces.
Across single-arm, dual-arm, humanoid, out-of-distribution, and real-world evaluations, the same formulation supports unified manipulation, cross-embodiment transfer, human-to-robot transfer, and data-efficient adaptation.
These results suggest that camera-centric action geometry provides a practical interface for scaling reusable robot policies across heterogeneous embodiments and demonstration sources.

\noindent \textbf{Limitations.}
\method{} relies on geometric information that must be estimated or calibrated reliably.
Errors in camera calibration, depth estimation, embodiment kinematics, or MediaPipe hand-keypoint localization can propagate into the camera-centric target and the downstream translator.
Cross-embodiment transfer remains challenging.
Although \method{} achieves non-zero ALOHA-to-ARX transfer, the gap from the source embodiment indicates that camera-centric motion alone does not eliminate morphology and kinematic mismatch.
Our real-world evaluation is intentionally controlled and limited in scale.
The experiments support human-to-robot transfer and data-efficient SFT adaptation, but broader real-world robustness requires evaluation across more robots, camera setups, object categories, and long-horizon tasks.

% \paragraph{Limitations}\method{} depends on reliable calibration, depth, kinematics, and hand-keypoint estimates, whose errors can propagate to control.
% Camera-centric motion reduces but does not eliminate morphology and kinematic mismatch, as indicated by the ALOHA-to-ARX gap.
% Our controlled real-world evaluation is limited in scale; broader validation across robots, cameras, objects, and long-horizon tasks remains future work.

\clearpage
\newcommand{\envelopeicon}{\ding{41}}

\section{Contributions and Acknowledgments}
\label{sec:contributions}

\noindent
\begin{minipage}[t]{0.47\linewidth}
  \subsubsection*{Core Contributors}
  \begin{itemize}[leftmargin=*, itemsep=0.15em, topsep=0.25em]
    \item Shaoqing Xu$^{*,\dagger}$
    \item Fang Li$^{*}$
    \item Guozhi Zhan$^{*}$
    \item Zhixiang Duan$^{*,\dagger}$
    \item Yuhan Wang
    \item Yuechen Luo
    \item Shengyin Jiang
    \item Hanbing Li
    \item Zhiying Du
    \item Hangjun Ye
    \item Zhi-xin Yang~\textsuperscript{\envelopeicon}
  \end{itemize}
\end{minipage}%
\hfill
\begin{minipage}[t]{0.47\linewidth}
  \subsubsection*{Contributors}
  \begin{itemize}[leftmargin=*, itemsep=0.15em, topsep=0.25em]
    \item Longlong Wang
    \item Longmei Jiang
    \item Weixiang Liang
    \item Ying Gong
    \item Yong Pan
    \item Ziping Zhao
    \item Zhiyuan Chen
    \item Yangwei You
    \item Kun Ma
    \item Qinyuan Liu
    
  \end{itemize}
\end{minipage}

\vspace{0.5em}
\noindent{\small\itshape
$^{*}$Equal core contribution.\quad
$^{\dagger}$Project Leads.\quad
\textsuperscript{\envelopeicon}Corresponding author.}

\subsubsection*{Acknowledgments}
\noindent
We thank the teams and maintainers of the models, datasets, benchmarks, simulation environments, and robotic platforms used in this work. These resources were used for research and evaluation purposes.

\clearpage
\bibliographystyle{plainnat}
\bibliography{main}

@inproceedings{brohan2023rt2,
  title = {RT-2: Vision-Language-Action Models Transfer Web Knowledge to Robotic Control},
  author = {Brohan, Anthony and Brown, Noah and Carbajal, Justice and Chebotar, Yevgen and Chen, Xi and Choromanski, Krzysztof and Ding, Tianli and Driess, Danny and Dubey, Avinava and Finn, Chelsea and Florence, Pete and Fu, Chuyuan and Arenas, Montse Gonzalez and Gopalakrishnan, Keerthana and Han, Kehang and Hausman, Karol and Herzog, Alexander and Hsu, Jasmine and Ichter, Brian and Irpan, Alex and Joshi, Nikhil and Julian, Ryan and Kalashnikov, Dmitry and Kuang, Yuheng and Leal, Isabel and Lee, Lisa and Lee, Tsang-Wei Edward and Levine, Sergey and Lu, Yao and Michalewski, Henryk and Mordatch, Igor and Pertsch, Karl and Rao, Kanishka and Reymann, Krista and Ryoo, Michael and Salazar, Grecia and Sanketi, Pannag and Sermanet, Pierre and Singh, Jaspiar and Singh, Anikait and Soricut, Radu and Tran, Huong and Vanhoucke, Vincent and Vuong, Quan and Wahid, Ayzaan and Welker, Stefan and Wohlhart, Paul and Wu, Jialin and Xia, Fei and Xiao, Ted and Xu, Peng and Xu, Sichun and Yu, Tianhe and Zitkovich, Brianna},
  booktitle = {Proceedings of The 7th Conference on Robot Learning},
  year = {2023}
}

@article{liu2024libero,
  title = {LIBERO: Benchmarking Knowledge Transfer for Lifelong Robot Learning},
  author = {Liu, Bo and Zhu, Yifeng and Gao, Chongkai and Feng, Yihao and Liu, Qiang and Zhu, Yuke and Stone, Peter},
  journal = {arXiv preprint arXiv:2306.03310},
  year = {2024}
}

@article{oneill2024openx,
  title = {Open X-Embodiment: Robotic Learning Datasets and RT-X Models},
  author = {{Open X-Embodiment Collaboration} and O'Neill, Abby and Rehman, Abdul and Gupta, Abhinav and Maddukuri, Abhiram and Gupta, Abhishek and Padalkar, Abhishek and Lee, Abraham and Pooley, Acorn and Gupta, Agrim and Mandlekar, Ajay and Jain, Ajinkya and Tung, Albert and Bewley, Alex and Herzog, Alexander and Irpan, Alex and Khazatsky, Alexander and Rai, Anant and Gupta, Anchit and Wang, Andrew and others},
  journal = {arXiv preprint arXiv:2310.08864},
  year = {2024}
}

@article{khazatsky2024droid,
  title = {DROID: A Large-Scale In-The-Wild Robot Manipulation Dataset},
  author = {Khazatsky, Alexander and Pertsch, Karl and Nair, Suraj and Balakrishna, Ashwin and Dasari, Sudeep and Karamcheti, Siddharth and Nasiriany, Soroush and Srirama, Mohan Kumar and Chen, Lawrence Yunliang and Ellis, Kirsty and others},
  journal = {arXiv preprint arXiv:2403.12945},
  year = {2024}
}

@article{kim2024openvla,
  title = {OpenVLA: An Open-Source Vision-Language-Action Model},
  author = {Kim, Moo Jin and Pertsch, Karl and Karamcheti, Siddharth and Xiao, Ted and Balakrishna, Ashwin and Nair, Suraj and Rafailov, Rafael and Foster, Ethan and Lam, Grace and Sanketi, Pannag and Vuong, Quan and Kollar, Thomas and Burchfiel, Benjamin and Tedrake, Russ and Sadigh, Dorsa and Levine, Sergey and Liang, Percy and Finn, Chelsea},
  journal = {arXiv preprint arXiv:2406.09246},
  year = {2024}
}

@article{hao2025mimo,
  title={Mimo-embodied: X-embodied foundation model technical report},
  author={Hao, Xiaoshuai and Zhou, Lei and Huang, Zhijian and Hou, Zhiwen and Tang, Yingbo and Zhang, Lingfeng and Li, Guang and Lu, Zheng and Ren, Shuhuai and Meng, Xianhui and others},
  journal={arXiv preprint arXiv:2511.16518},
  year={2025}
}

@article{yang2026fpc,
  title={Fpc-vla: A vision-language-action framework with a supervisor for failure prediction and correction},
  author={Yang, Yifan and Duan, Zhixiang and Xie, Tianshi and Cao, Fuyu and Shen, Pinxi and Song, Peili and Zhao, Chenyang and Jin, Piaopiao and Sun, Guokang and Xu, Shaoqing and others},
  journal={Expert Systems with Applications},
  pages={131742},
  year={2026},
  publisher={Elsevier}
}

@article{li2026spaact,
  title={SpaAct: Spatially-Activated Transition Learning with Curriculum Adaptation for Vision-Language Navigation},
  author={Li, Pengna and Wu, Kangyi and Xu, Shaoqing and Li, Fang and Li, Hanbing and Zhao, Lin and Lyu, Kailin and Chen, Long and Yang, Zhi-Xin and Zheng, Nanning},
  journal={arXiv preprint arXiv:2604.27620},
  year={2026}
}

@inproceedings{li2026think,
  title={Think before Go: Hierarchical Reasoning for Image-goal Navigation},
  author={Li, Pengna and Wu, Kangyi and Xu, Shaoqing and Li, Fang and Zhao, Lin and Chen, Long and Yang, Zhi-Xin and Zheng, Nanning},
  booktitle={Proceedings of the 64th Annual Meeting of the Association for Computational Linguistics (Volume 1: Long Papers)},
  pages={29336--29357},
  year={2026}
}

@article{lu2026xiaomi,
  title={Xiaomi OneVL: One-Step Latent Reasoning and Planning with Vision-Language Explanation},
  author={Lu, Jinghui and Guan, Jiayi and Huang, Zhijian and Li, Jinlong and Li, Guang and Kong, Lingdong and Li, Yingyan and Wang, Han and Xu, Shaoqing and Luo, Yuechen and others},
  journal={arXiv preprint arXiv:2604.18486},
  year={2026}
}

@article{luo2026last,
  title={Last-vla: Thinking in latent spatio-temporal space for vision-language-action in autonomous driving},
  author={Luo, Yuechen and Li, Fang and Xu, Shaoqing and Ji, Yang and Zhang, Zehan and Wang, Bing and Shen, Yuannan and Cui, Jianwei and Chen, Long and Chen, Guang and others},
  journal={arXiv preprint arXiv:2603.01928},
  year={2026}
}

@inproceedings{chen2026vilta,
  title={VILTA: A VLM-in-the-loop adversary for enhancing driving policy robustness},
  author={Chen, Qimao and Li, Fang and Xu, Shaoqing and Lai, Zhiyi and Xie, Zixun and Luo, Yuechen and Jiang, Shengyin and Li, Hanbing and Chen, Long and Wang, Bing and others},
  booktitle={Proceedings of the AAAI Conference on Artificial Intelligence},
  volume={40},
  number={4},
  pages={2984--2992},
  year={2026}
}

@inproceedings{luo2026unleashing,
  title={Unleashing vla potentials in autonomous driving via explicit learning from failures},
  author={Luo, Yuechen and Li, Fang and Chen, Qimao and Xu, Shaoqing and Liu, Jiaxin and Song, Ziying and Yang, Zhi-xin and Wen, Fuxi},
  booktitle={Proceedings of the IEEE/CVF Conference on Computer Vision and Pattern Recognition},
  pages={24833--24842},
  year={2026}
}

@article{luo2025adathinkdrive,
  title={Adathinkdrive: Adaptive thinking via reinforcement learning for autonomous driving},
  author={Luo, Yuechen and Li, Fang and Xu, Shaoqing and Lai, Zhiyi and Yang, Lei and Chen, Qimao and Luo, Ziang and Xie, Zixun and Jiang, Shengyin and Liu, Jiaxin and others},
  journal={arXiv preprint arXiv:2509.13769},
  year={2025}
}

@article{black2024pi0,
  title = {$\pi_0$: A Vision-Language-Action Flow Model for General Robot Control},
  author = {Black, Kevin and Brown, Noah and Driess, Danny and Esmail, Adnan and Equi, Michael and Finn, Chelsea and Fusai, Niccolo and Groom, Lachy and Hausman, Karol and Ichter, Brian and Jakubczak, Szymon and Jones, Tim and Ke, Liyiming and Levine, Sergey and Li-Bell, Adrian and Mothukuri, Mohith and Nair, Suraj and Pertsch, Karl and Shi, Lucy Xiaoyang and Tanner, James and Vuong, Quan and Walling, Anna and Wang, Haohuan and Zhilinsky, Ury},
  journal = {arXiv preprint arXiv:2410.24164},
  year = {2024}
}

@article{song2025hume,
  title={Hume: Introducing system-2 thinking in visual-language-action model},
  author={Song, Haoming and Qu, Delin and Yao, Yuanqi and Chen, Qizhi and Lv, Qi and Tang, Yiwen and Shi, Modi and Ren, Guanghui and Yao, Maoqing and Zhao, Bin and others},
  journal={arXiv preprint arXiv:2505.21432},
  year={2025}
}

@article{physicalintelligence2025pi05,
  title = {$\pi_{0.5}$: A Vision-Language-Action Model with Open-World Generalization},
  author = {{Physical Intelligence} and Black, Kevin and Brown, Noah and Darpinian, James and Dhabalia, Karan and Driess, Danny and Esmail, Adnan and Equi, Michael and Finn, Chelsea and Fusai, Niccolo and Galliker, Manuel Y. and Ghosh, Dibya and Groom, Lachy and Hausman, Karol and Ichter, Brian and Jakubczak, Szymon and Jones, Tim and Ke, Liyiming and LeBlanc, Devin and Levine, Sergey and Li-Bell, Adrian and Mothukuri, Mohith and Nair, Suraj and Pertsch, Karl and Ren, Allen Z. and Shi, Lucy Xiaoyang and Smith, Laura and Springenberg, Jost Tobias and Stachowicz, Kyle and Tanner, James and Vuong, Quan and Walke, Homer and Walling, Anna and Wang, Haohuan and Yu, Lili and Zhilinsky, Ury},
  journal = {arXiv preprint arXiv:2504.16054},
  year = {2025}
}

@article{black2025fast,
  title = {{FAST}: Efficient Action Tokenization for Vision-Language-Action Models},
  author = {Pertsch, Karl and Stachowicz, Kyle and Ichter, Brian and Driess, Danny and Nair, Suraj and Vuong, Quan and Mees, Oier and Finn, Chelsea and Levine, Sergey},
  year = {2025},
  journal = {arXiv preprint arXiv:2501.09747}
}

@article{mu2024robotwin,
  title = {RoboTwin: Dual-Arm Robot Benchmark with Generative Digital Twins (early version)},
  author = {Mu, Yao and Chen, Tianxing and Peng, Shijia and Chen, Zanxin and Gao, Zeyu and Zou, Yude and Lin, Lunkai and Xie, Zhiqiang and Luo, Ping},
  journal = {arXiv preprint arXiv:2409.02920},
  year = {2024}
}

@article{nasiriany2024robocasa,
  title = {RoboCasa: Large-Scale Simulation of Everyday Tasks for Generalist Robots},
  author = {Nasiriany, Soroush and Maddukuri, Abhiram and Zhang, Lance and Parikh, Adeet and Lo, Aaron and Joshi, Abhishek and Mandlekar, Ajay and Zhu, Yuke},
  journal = {arXiv preprint arXiv:2406.02523},
  year = {2024}
}

@article{bharadhwaj2024gen2act,
  title = {Gen2Act: Human Video Generation in Novel Scenarios Enables Generalizable Robot Manipulation},
  author = {Bharadhwaj, Homanga and Vakil, Jatin and Sharma, Mohit and Gupta, Abhinav and Tulsiani, Shubham and Kumar, Vikash},
  journal = {arXiv preprint arXiv:2409.16283},
  year = {2024}
}

@article{chen2026videomanip,
  title = {VIDEOMANIP: Dexterous Manipulation Policies from RGB Human Videos via 4D Hand-Object Trajectory Reconstruction},
  author = {Chen, Sirui and Chen, Yanjie Ze and Li, Zhe and Black, Michael J. and Jampani, Varun and Fazeli, Nima and Fan, Linxi},
  journal = {arXiv preprint arXiv:2602.09013},
  year = {2026}
}

@article{fang2026molmoact2,
  title = {MolmoAct2: Action Reasoning Models for Real-world Deployment},
  author = {Fang, Haoquan and Duan, Jiafei and Clay, Donovan and Wang, Sam and Liu, Shuo and Huang, Weikai and Fan, Xiang and Tsai, Wei-Chuan and Chen, Shirui and Wang, Yi Ru and Xing, Shanli and Cho, Jaemin and Park, Jae Sung and Eftekhar, Ainaz and Sushko, Peter and Farley, Karen and Wadhwa, Angad and Harrison, Cole and Han, Winson and Lee, Ying-Chun and VanderBilt, Eli and Hendrix, Rose and Ellawela, Suveen and Ngoo, Lucas and Chai, Joyce and Ren, Zhongzheng and Farhadi, Ali and Fox, Dieter and Krishna, Ranjay},
  journal = {arXiv preprint arXiv:2605.02881},
  year = {2026}
}

@article{wang2026qwenvla,
  title = {Qwen-VLA: Unifying Vision-Language-Action Modeling across Tasks, Environments, and Robot Embodiments},
  author = {Wang, Qiuyue and Li, Mingsheng and Guan, Jian and Ye, Jinhui and Xie, Sicheng and Liu, Yitao and Chen, Junhao and Liang, Zhixuan and Zhang, Jie and Hu, Xintong and Huang, Xuhong and Lin, Pei and Lin, Junyang and Liu, Dayiheng and Bai, Shuai and Zhou, Jingren and Zhang, Jiazhao and Yuan, Haoqi and Zhou, Gengze and Yin, Hang and Wang, Ye and Huang, Yiyang and Lei, Zixing and Peng, Wujian and Chen, Delin and Zheng, Yingming and Fan, Jingyang and Zhuang, Xianwei and Zhou, Xin and Li, Haoyang and Chen, Anzhe and Zhang, Tong and Liu, Xuejing and Sun, Yuchong and Chen, Ruizhe and Li, Zhaohai and Lu, Chenxu and Yang, Zhibo and Yu, Tao and Chen, Xionghui},
  journal = {arXiv preprint arXiv:2605.30280},
  year = {2026}
}

@article{octo2024,
  title = {{Octo}: An Open-Source Generalist Robot Policy},
  author = {{Octo Model Team} and Ghosh, Dibya and Walke, Homer and Pertsch, Karl and Black, Kevin and Mees, Oier and Dasari, Sudeep and Hejna, Joey and Kreiman, Tobias and Xu, Charles and others},
  journal = {arXiv preprint arXiv:2405.12213},
  year = {2024}
}

@article{bjorck2025groot,
  title = {{GR00T N1}: An Open Foundation Model for Generalist Humanoid Robots},
  author = {{NVIDIA} and Bjorck, Johan and Castaneda, Fernando and Cherniadev, Nikita and Da, Xingye and Ding, Runyu and Fan, Linxi and Fang, Yu and Fox, Dieter and Hu, Fengyuan and others},
  journal = {arXiv preprint arXiv:2503.14734},
  year = {2025}
}

@article{liu2024rdt,
  title = {{RDT-1B}: A Diffusion Foundation Model for Bimanual Manipulation},
  author = {Liu, Songming and Wu, Lingxuan and Li, Bangguo and Tan, Hengkai and Chen, Huayu and Wang, Zhengyi and Xu, Ke and Su, Hang and Zhu, Jun},
  journal = {arXiv preprint arXiv:2410.07864},
  year = {2024}
}

@article{kim2025openvlaoft,
  title = {Fine-Tuning Vision-Language-Action Models: Optimizing Speed and Success},
  author = {Kim, Moo Jin and Finn, Chelsea and Liang, Percy},
  journal = {arXiv preprint arXiv:2502.19645},
  year = {2025}
}

@article{qu2025eo,
  title={EO-1: An Open Unified Embodied Foundation Model for General Robot Control},
  author={Qu, Delin and Song, Haoming and Chen, Qizhi and Chen, Zhaoqing and Gao, Xianqiang and Wang, Dong and Ye, Xinyi and Lv, Qi and Shi, Modi and Ren, Guanghui and others},
  journal={arXiv preprint arXiv:2508.21112},
  year={2025}
}

@article{hung2025nora,
  title = {{NORA}: A Small Open-Sourced Generalist Vision Language Action Model for Embodied Tasks},
  author = {Hung, Chia-Yu and Sun, Qi and Hong, Pengfei and Zadeh, Amir and Li, Chuan and Tan, U and Majumder, Navonil and Poria, Soujanya and others},
  journal = {arXiv preprint arXiv:2504.19854},
  year = {2025}
}

@article{qu2025spatialvla,
  title = {{SpatialVLA}: Exploring Spatial Representations for Visual-Language-Action Model},
  author = {Qu, Delin and Song, Haoming and Chen, Qizhi and Yao, Yuanqi and Ye, Xinyi and Ding, Yan and Wang, Zhigang and Gu, JiaYuan and Zhao, Bin and Wang, Dong and Li, Xuelong},
  journal = {arXiv preprint arXiv:2501.15830},
  year = {2025}
}

@article{zhao2025cotvla,
  title = {{CoT-VLA}: Visual Chain-of-Thought Reasoning for Vision-Language-Action Models},
  author = {Zhao, Qingqing and Lu, Yao and Kim, Moo Jin and Fu, Zipeng and Zhang, Zhuoyang and Wu, Yecheng and Li, Zhaoshuo and Ma, Qianli and Han, Song and Finn, Chelsea and Handa, Ankur and Liu, Ming-Yu and Xiang, Donglai and Wetzstein, Gordon and Lin, Tsung-Yi},
  journal = {arXiv preprint arXiv:2503.22020},
  year = {2025}
}

@article{zheng2025xvla,
  title = {{X-VLA}: Soft-Prompted Transformer as Scalable Cross-Embodiment Vision-Language-Action Model},
  author = {Zheng, Jinliang and Li, Jianxiong and Wang, Zhihao and Liu, Dongxiu and Kang, Xirui and Feng, Yuchun and Zheng, Yinan and Zou, Jiayin and Chen, Yilun and Zeng, Jia and Zhang, Ya-Qin and Pang, Jiangmiao and Liu, Jingjing and Wang, Tai and Zhan, Xianyuan},
  journal = {arXiv preprint arXiv:2510.10274},
  year = {2025}
}

@article{cen2025worldvla,
  title = {{WorldVLA}: Towards Autoregressive Action World Model},
  author = {Cen, Jun and Yu, Chaohui and Yuan, Hangjie and Jiang, Yuming and Huang, Siteng and Guo, Jiayan and Li, Xin and Song, Yibing and Luo, Hao and Wang, Fan and others},
  journal = {arXiv preprint arXiv:2506.21539},
  year = {2025}
}

@article{bu2025univla,
  title = {{UniVLA}: Learning to Act Anywhere with Task-Centric Latent Actions},
  author = {Bu, Qingwen and Yang, Yanting and Cai, Jisong and Gao, Shenyuan and Ren, Guanghui and Yao, Maoqing and Luo, Ping and Li, Hongyang},
  journal = {Robotics: Science and Systems},
  year = {2025}
}

@article{tan2025interactive,
  title = {Interactive Post-Training for Vision-Language-Action Models},
  author = {Tan, Shuhan and Dou, Kairan and Zhao, Yue and Kr"ahenb"uhl, Philipp},
  journal = {arXiv preprint arXiv:2505.17016},
  year = {2025}
}

@article{nair2022r3m,
  title = {{R3M}: A Universal Visual Representation for Robot Manipulation},
  author = {Nair, Suraj and Rajeswaran, Aravind and Kumar, Vikash and Finn, Chelsea and Gupta, Abhinav},
  journal = {arXiv preprint arXiv:2203.12601},
  year = {2022}
}

@article{qin2021dexmv,
  title = {{DexMV}: Imitation Learning for Dexterous Manipulation from Human Videos},
  author = {Qin, Yuzhe and Wu, Yueh-Hua and Liu, Shaowei and Jiang, Hanwen and Yang, Ruihan and Fu, Yang and Wang, Xiaolong},
  journal = {arXiv preprint arXiv:2108.05877},
  year = {2021}
}

@article{bahl2022whirl,
  title = {Human-to-Robot Imitation in the Wild},
  author = {Bahl, Shikhar and Gupta, Abhinav and Pathak, Deepak},
  journal = {arXiv preprint arXiv:2207.09450},
  year = {2022}
}

@article{mandlekar2021robomimic,
  title = {What Matters in Learning from Offline Human Demonstrations for Robot Manipulation},
  author = {Mandlekar, Ajay and Xu, Danfei and Wong, Josiah and Nasiriany, Soroush and Wang, Chen and Kulkarni, Rohun and Fei-Fei, Li and Savarese, Silvio and Zhu, Yuke and Martin-Martin, Roberto},
  journal = {arXiv preprint arXiv:2108.03298},
  year = {2021}
}

@article{zhao2023aloha,
  title = {Learning Fine-Grained Bimanual Manipulation with Low-Cost Hardware},
  author = {Zhao, Tony Z. and Kumar, Vikash and Levine, Sergey and Finn, Chelsea},
  journal = {arXiv preprint arXiv:2304.13705},
  year = {2023}
}

@article{chi2023diffusionpolicy,
  title = {Diffusion Policy: Visuomotor Policy Learning via Action Diffusion},
  author = {Chi, Cheng and Xu, Zhenjia and Feng, Siyuan and Cousineau, Eric and Du, Yilun and Burchfiel, Benjamin and Tedrake, Russ and Song, Shuran},
  journal = {arXiv preprint arXiv:2303.04137},
  year = {2023}
}

@article{zeng2020transporter,
  title = {Transporter Networks: Rearranging the Visual World for Robotic Manipulation},
  author = {Zeng, Andy and Florence, Pete and Tompson, Jonathan and Welker, Stefan and Chien, Jonathan and Attarian, Maria and Armstrong, Travis and Krasin, Ivan and Duong, Dan and Wahid, Ayzaan and others},
  journal = {arXiv preprint arXiv:2010.14406},
  year = {2020}
}

@article{shridhar2022peract,
  title = {Perceiver-Actor: A Multi-Task Transformer for Robotic Manipulation},
  author = {Shridhar, Mohit and Manuelli, Lucas and Fox, Dieter},
  journal = {arXiv preprint arXiv:2209.05451},
  year = {2022}
}

@article{goyal2023rvt,
  title = {{RVT}: Robotic View Transformer for 3D Object Manipulation},
  author = {Goyal, Ankit and Xu, Jie and Guo, Yijie and Blukis, Valts and Chao, Yu-Wei and Fox, Dieter},
  journal = {arXiv preprint arXiv:2306.14896},
  year = {2023}
}

@article{wu2026lingbotvla,
  title = {A Pragmatic {VLA} Foundation Model},
  author = {Wu, Wei and Lu, Fan and Wang, Yunnan and Yang, Shuai and Liu, Shi and Wang, Fangjing and Zhu, Qian and Sun, He and Wang, Yong and Ma, Shuailei and others},
  journal = {arXiv preprint arXiv:2601.18692},
  year = {2026}
}

@article{wu2024robomind,
  title = {{RoboMIND}: Benchmark on Multi-embodiment Intelligence Normative Data for Robot Manipulation},
  author = {Wu, Kun and Hou, Chengkai and Liu, Jiaming and Che, Zhengping and Ju, Xiaozhu and Yang, Zhuqin and Li, Meng and Zhao, Yinuo and Xu, Zhiyuan and Yang, Guang and others},
  journal = {arXiv preprint arXiv:2412.13877},
  year = {2024}
}

@article{wang2024crosslatent,
  title = {Cross-Embodiment Robot Manipulation Skill Transfer using Latent Space Alignment},
  author = {Wang, Tianyu and Bhatt, Dwait and Wang, Xiaolong and Atanasov, Nikolay},
  journal = {arXiv preprint arXiv:2406.01968},
  year = {2024}
}

@article{yan2026unifiedlatent,
  title = {Learning a Unified Latent Space for Cross-Embodiment Robot Control},
  author = {Yan, Yashuai and Lee, Dongheui},
  journal = {arXiv preprint arXiv:2601.15419},
  year = {2026}
}

@article{yu2025egomi,
  title = {{EgoMI}: Learning Active Vision and Whole-Body Manipulation from Egocentric Human Demonstrations},
  author = {Yu, Justin and Shentu, Yide and Wu, Di and Abbeel, Pieter and Goldberg, Ken and Wu, Philipp},
  journal = {arXiv preprint arXiv:2511.00153},
  year = {2025}
}

@article{shi2026egohumanoid,
  title = {{EgoHumanoid}: Unlocking In-the-Wild Loco-Manipulation with Robot-Free Egocentric Demonstration},
  author = {Shi, Modi and Peng, Shijia and Chen, Jin and Jiang, Haoran and Li, Yinghui and Huang, Di and Luo, Ping and Li, Hongyang and Chen, Li},
  journal = {arXiv preprint arXiv:2602.10106},
  year = {2026}
}

@article{bai2025qwen3vl,
  title = {{Qwen3-VL} Technical Report},
  author = {Bai, Shuai and Cai, Yuxuan and Chen, Ruizhe and Chen, Keqin and Chen, Xionghui and Cheng, Zesen and Deng, Lianghao and Ding, Wei and Gao, Chang and Ge, Chunjiang and others},
  journal = {arXiv preprint arXiv:2511.21631},
  year = {2025}
}

@article{starvla2026,
  title = {{StarVLA}: A Lego-like Codebase for Vision-Language-Action Model Developing},
  author = {{StarVLA Community}},
  journal = {arXiv preprint arXiv:2604.05014},
  year = {2026}
}

@article{li2026aceego,
  title = {{ACE-Ego-0}: Unifying Egocentric Human and Robotic Data for {VLA} Pretraining},
  author = {Li, Hao and Zhao, Ganlong and Liu, Yufei and Hou, Haotian and Ye, Guoquan and Fang, Tongyan and Liu, Chunxiao and Huang, Siyuan and Liu, Jianbo and Wang, Xiaogang and Li, Hongsheng},
  journal = {arXiv preprint arXiv:2606.17200},
  year = {2026}
}

@article{qwen2026robotmanip,
  title = {{Qwen-RobotManip} Technical Report: Alignment Unlocks Scale for Robotic Manipulation Foundation Models},
  author = {{Qwen Team}},
  journal = {arXiv preprint arXiv:2606.17846},
  year = {2026}
}

@article{yang2026abotm0,
  title = {{ABot-M0}: {VLA} Foundation Model for Robotic Manipulation with Action Manifold Learning},
  author = {Yang, Yandan and Zeng, Shuang and Lin, Tong and Chang, Xinyuan and Qi, Dekang and Xiao, Junjin and Liu, Haoyun and Chen, Ronghan and Chen, Yuzhi and Huo, Dongjie and Xiong, Feng and Wei, Xing and Ma, Zhiheng and Xu, Mu},
  journal = {arXiv preprint arXiv:2602.11236},
  year = {2026}
}

@article{luo2026beingh05,
  title = {{Being-H0.5}: Scaling Human-Centric Robot Learning for Cross-Embodiment Generalization},
  author = {Luo, Hao and Wang, Ye and Zhang, Wanpeng and Zheng, Sipeng and Xi, Ziheng and Xu, Chaoyi and Xu, Haiweng and Yuan, Haoqi and Zhang, Chi and Wang, Yiqing and Feng, Yicheng and Lu, Zongqing},
  journal = {arXiv preprint arXiv:2601.12993},
  year = {2026}
}

@article{luo2026beingh07,
  title = {{Being-H0.7}: A Latent World-Action Model from Egocentric Videos},
  author = {Luo, Hao and Zhang, Wanpeng and Feng, Yicheng and Zheng, Sipeng and Xu, Haiweng and Xu, Chaoyi and Xi, Ziheng and Fu, Yuhui and Lu, Zongqing},
  journal = {arXiv preprint arXiv:2605.00078},
  year = {2026}
}

@article{li2026zr0,
  title = {Training Vision-Language-Action Models with Dense Embodied Chain-of-Thought Supervision},
  author = {Li, Haoyang and Li, Guanlin and Feng, Youhe and Zhao, Chen and Wang, Zhuoran and Li, Yang and Wei, Qizhe and Bao, Shifeng and Shen, Haitao and Zhao, Yihan and Yang, Tong and Zhang, Jing},
  journal = {arXiv preprint arXiv:2606.30552},
  year = {2026}
}

@article{su2026demavla,
  title = {{DeMaVLA}: A Vision-Language-Action Foundation Model for Generalizable Deformable Manipulation},
  author = {Su, Taiyi and Zhu, Jian and Wang, Tianjian and He, Youzhang and Huang, Zitai and Zhang, Jianjun and Ma, Chong and Wang, Hanyang and Zhang, Tianjiao and Yin, Munan and Ding, Weihao and Xu, Yi},
  journal = {arXiv preprint arXiv:2605.31286},
  year = {2026}
}

@article{chen2026oasis,
  title = {{OASIS}: Observation-Action Space Alignment via {SE(3)} Trajectory Prediction for Robotic Manipulation},
  author = {Chen, Xinzhe and Ren, Sihua and Huang, Liqi and Sun, Haowen and Li, Mingyang and Chen, Xingyu and Liu, Zeyang and Lan, Xuguang},
  journal = {arXiv preprint arXiv:2605.25829},
  year = {2026}
}

@article{zhang2026joyaira,
  title = {{JoyAI-RA 0.1}: A Foundation Model for Robotic Autonomy},
  author = {Zhang, Tianle and Yuan, Zhihao and Chi, Dafeng and Liu, Peidong and Li, Dongwei and Hu, Kejun and Zhang, Likui and Nie, Junnan and Wei, Ziming and Chen, Zengjue and others},
  journal = {arXiv preprint arXiv:2604.20100},
  year = {2026}
}

@article{Lugaresi2019MediaPipe,
  title = {{MediaPipe}: A Framework for Building Perception Pipelines},
  author = {Lugaresi, Camillo and Tang, Jiuqiang and Nash, Hadon and McClanahan, Chris and Uboweja, Esha and Hays, Michael and Zhang, Fan and Chang, Chuo-Ling and Yong, Ming Guang and Lee, Juhyun and Chang, Wan-Teh and Hua, Wei and Georg, Manfred and Grundmann, Matthias},
  journal = {arXiv preprint arXiv:1906.08172},
  year = {2019}
}

@inproceedings{ji2025robobrain,
  title = {{RoboBrain}: A Unified Brain Model for Robotic Manipulation from Abstract to Concrete},
  author = {Ji, Yuheng and Li, Yehui and Wang, Yiyang and Luo, Qingfeng and Liu, Zhaolei and Hu, Boyong and Chen, Xiyao and Zang, Yuhang and Zhang, Chi and Li, Xiaomeng and others},
  booktitle = {Proceedings of the IEEE/CVF Conference on Computer Vision and Pattern Recognition},
  year = {2025}
}

@article{yakefu2025robochallenge,
  title = {{RoboChallenge}: Large-scale Real-robot Evaluation of Embodied Policies},
  author = {Yakefu, Shuaidi and others},
  journal = {arXiv preprint arXiv:2510.17950},
  year = {2025}
}

@article{wu2025robocoin,
  title = {{RoboCOIN}: An Open-Sourced Bimanual Robotic Data Collection for Integrated Manipulation},
  author = {Wu, Shihan and others},
  journal = {arXiv preprint arXiv:2511.17441},
  year = {2025}
}

@article{tian2025interndataa1,
  title = {{InternData-A1}: Pioneering High-Fidelity Synthetic Data for Pre-training Generalist Policy},
  author = {Tian, Yang and Yang, Yuyin and Xie, Yiman and Cai, Zetao and Shi, Xu and Gao, Ning and Liu, Hangxu and Jiang, Xuekun and Qiu, Zherui and Yuan, Feng and Li, Yaping and Wang, Ping and Cai, Junhao and Zeng, Jia and Dong, Hao and Pang, Jiangmiao},
  journal = {arXiv preprint arXiv:2511.16651},
  year = {2025}
}

@article{vitra,
  title = {Scalable Vision-Language-Action Model Pretraining for Robotic Manipulation with Real-Life Human Activity Videos},
  author = {Li, Qixiu and Deng, Yu and Liang, Yaobo and Luo, Lin and Zhou, Lei and Yao, Chengtang and Zeng, Lingqi and Feng, Zhiyuan and Liang, Huizhi and Xu, Sicheng and others},
  journal = {arXiv preprint arXiv:2510.21571},
  year = {2025}
}

@article{egodex,
  title = {{EgoDex}: Learning Dexterous Manipulation from Large-Scale Egocentric Video},
  author = {Hoque, Ryan and Huang, Peide and Yoon, David J. and Sivapurapu, Mouli and Zhang, Jian},
  journal = {arXiv preprint arXiv:2505.11709},
  year = {2025}
}

@article{egoverse,
  title = {{EgoVerse}: An Egocentric Human Dataset for Robot Learning from Around the World},
  author = {Punamiya, Ryan and Kareer, Simar and Liu, Zeyi and Citron, Josh and Qiu, Ri-Zhao and Cai, Xiongyi and Gavryushin, Alexey and Chen, Jiaqi and Liconti, Davide and Zhu, Lawrence Y. and others},
  journal = {arXiv preprint arXiv:2604.07607},
  year = {2026}
}

@article{zhou2025roborefer,
  title = {{RoboRefer}: Towards Spatial Referring with Reasoning in Vision-Language Models for Robotics},
  author = {Zhou, Enshen and An, Jingkun and Chi, Cheng and Han, Yi and Rong, Shanyu and Zhang, Chi and Wang, Pengwei and Wang, Zhongyuan and Huang, Tie-Jun and Sheng, Lu and Zhang, Shanghang},
  journal = {arXiv preprint arXiv:2506.04308},
  year = {2025}
}

@inproceedings{sermanet2024robovqa,
  title = {{RoboVQA}: Multimodal Long-Horizon Reasoning for Robotics},
  author = {Sermanet, Pierre and others},
  booktitle = {Proceedings of the IEEE International Conference on Robotics and Automation},
  year = {2024}
}

@inproceedings{tang2025roboafford,
  title = {{RoboAfford}: A Dataset and Benchmark for Enhancing Object and Spatial Affordance Learning in Robot Manipulation},
  author = {Tang, Yingbo and others},
  booktitle = {Proceedings of the ACM International Conference on Multimedia},
  year = {2025}
}

@article{wang2026vpvla,
  title = {{VP-VLA}: Visual Prompting as an Interface for Vision-Language-Action Models},
  author = {Wang, Zixuan and Chen, Yuxin and Liu, Yuqi and Ye, Jinhui and Chen, Pengguang and Lu, Changsheng and Liu, Shu and Yu, Bei and Jia, Jiaya},
  journal = {arXiv preprint arXiv:2603.22003},
  year = {2026}
}

@inproceedings{zhou2019continuity,
  title={On the continuity of rotation representations in neural networks},
  author={Zhou, Yi and Barnes, Connelly and Lu, Jingwan and Yang, Jimei and Li, Hao},
  booktitle={2019 IEEE/CVF Conference on Computer Vision and Pattern Recognition (CVPR)},
  pages={5738--5746},
  year={2019},
  organization={IEEE}
}

\clearpage
\beginappendix
\counterwithin{table}{section}
\counterwithin{figure}{section}

\noindent\textbf{Overview.}
This supplementary material provides additional details on the architecture, evaluation, and qualitative analysis of \method{}. \Cref{appendix:training-evaluation-details} presents the training schedule, head architectures, sparse command layout, evaluation protocols, and question--answer examples (\Cref{app:training-schedule,app:head-architectures,app:sparse-command-layout,sec:real-world-protocol,app:qa-examples}). \Cref{app:extended-results} reports transfer results, pooling ablations, and benchmark-level breakdowns (\Cref{sec:generalization-transfer,sec:pooling-ablation-appendix,sec:libero-suite-appendix,sec:robocasa-gr1-appendix,sec:robotwin-appendix}). \Cref{appendix-vis} provides human-to-robot and qualitative visualizations (\Cref{app:human-transfer-details,app:human_demonstration,app:qualitative-examples}), while \Cref{appendix:failure-analysis} discusses representative failure cases and limitations.

\section{Training \& Evaluation Details}
\label{appendix:training-evaluation-details}

This section records the optimization, control-layout, and evaluation details omitted from the main paper.

\subsection{Training Schedule and Optimization}
\label{app:training-schedule}

\begin{table*}[ht]
\centering
% \small
\setlength{\tabcolsep}{3pt}
\small
\caption{Stage-wise \method{} training schedule. Batch size denotes the per-device batch size; global batch size is computed before any gradient accumulation.}
\label{tab:training-schedule-appendix}
\begin{tabular}{lcccccc}
\toprule
Stage & Objective & GPUs & Steps & Image size & Batch / GPU & Global batch \\
\midrule
1 & Camera-centric specialization & 128 H20 & 200K & $224\times224$ & 12 & 1,536 \\
2 & Geometry-conditioned action translation & 8 H20 & 10K & N/A & 128 & 1,024 \\
3 & Joint robot-human training & 64 H20 & 10K & $224\times224$ & 12 & 768 \\
\bottomrule
\end{tabular}
\end{table*}

\Cref{tab:training-schedule-appendix} summarizes the stage-wise compute and optimization schedule used to train \method{}.
Stages 1 and 3 resize observation images to $224\times224$ and use an action chunk horizon of 30; Stage 2 trains the geometry-conditioned translator from trajectory, state, and calibration inputs without image observations.
Unless otherwise noted, we use AdamW with $\beta_1=0.9$, $\beta_2=0.95$, $\epsilon=10^{-8}$, gradient clipping at 1.0, gradient accumulation of 1, and cosine decay with a minimum learning rate of $5\times10^{-7}$ after 1,000 warmup steps.
We use separate learning rates for the pre-trained VLM components and the action module: $1\times10^{-5}$ for the base backbone and Qwen-VL interface, and $1\times10^{-4}$ for the action model.

\subsection{Motion and Action Head Architectures}
\label{app:head-architectures}

\begin{figure}[!ht]
\centering
\includegraphics[width=\linewidth]{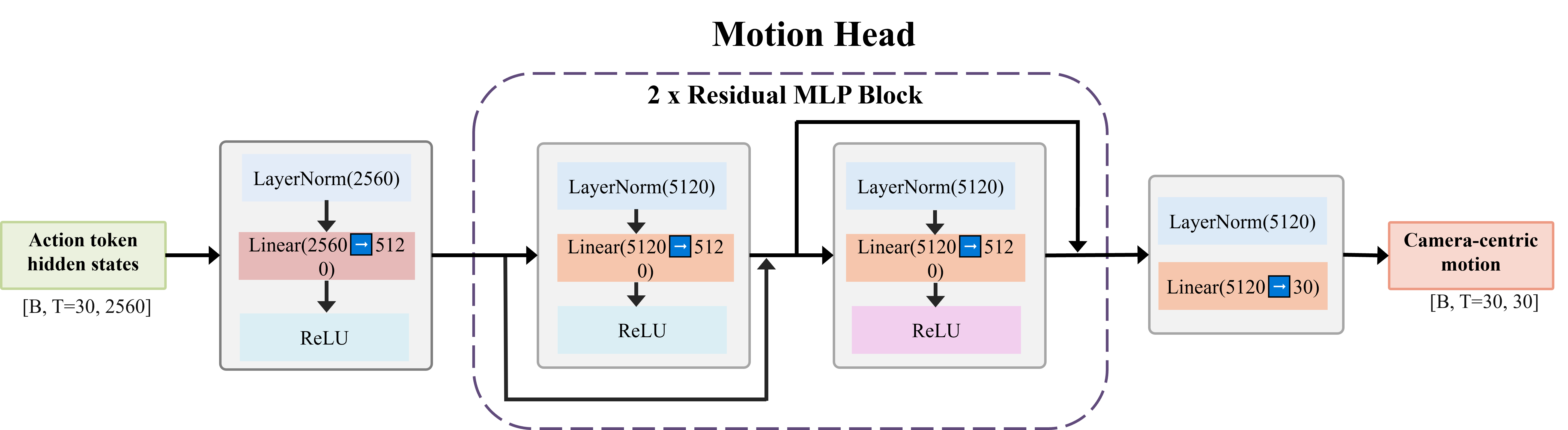}
\caption{Architecture of the motion head. Two residual MLP blocks map VLM action-token features to a 30-step camera-centric motion prediction.}
\label{fig:motion-head-architecture}
\end{figure}

\begin{figure}[!t]
\centering
\includegraphics[width=\linewidth]{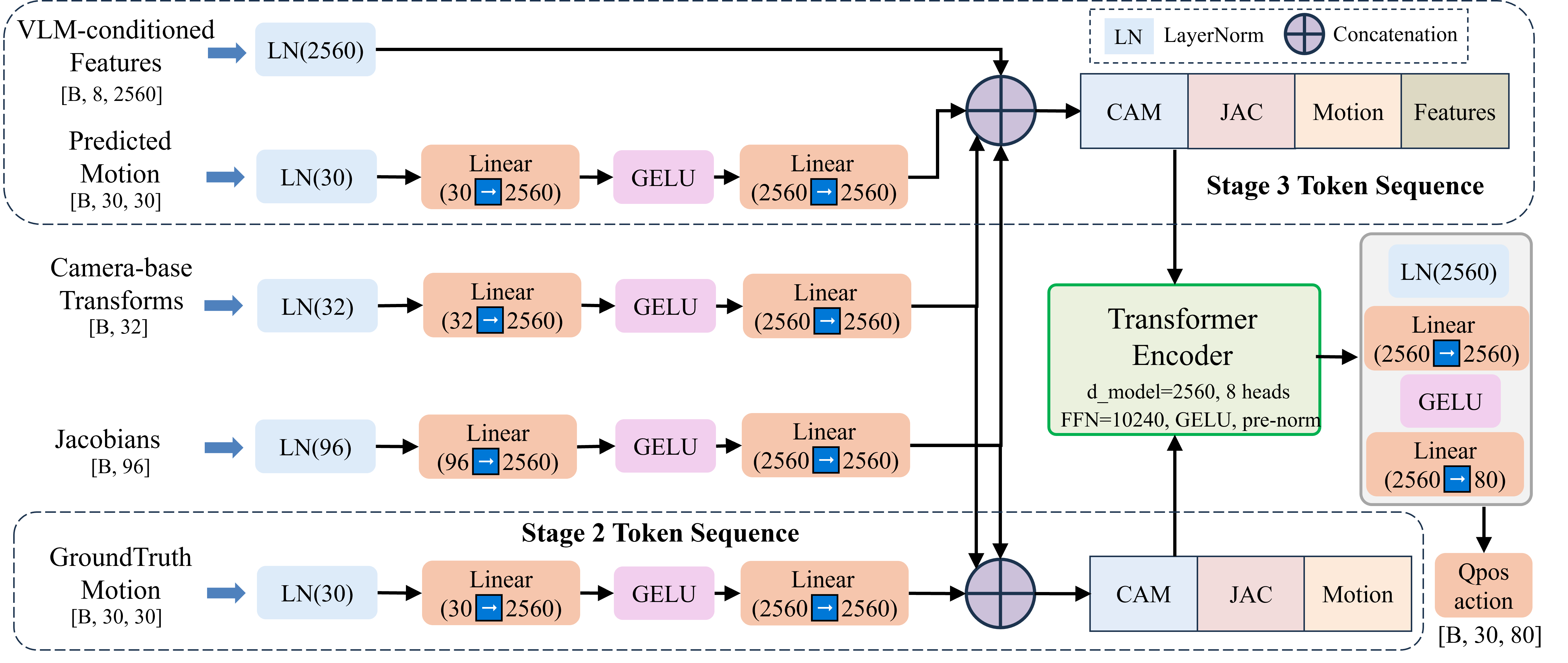}
\caption{Architecture of the geometry-conditioned action head. VLM-conditioned features, camera pose, Jacobian, and ground-truth or predicted motion are fused by a Transformer encoder to produce a 30-step qpos command chunk.}
\label{fig:action-head-architecture}
\end{figure}

\Cref{fig:motion-head-architecture,fig:action-head-architecture} detail the two prediction heads used by \method{}. The motion head first maps the VLM action-token features to a shared camera-centric motion representation, after which the action head combines this representation with embodiment geometry to predict executable commands.

\paragraph{Motion head}
As shown in \Cref{fig:motion-head-architecture}, the motion head projects each 2,560-dimensional action-token feature to a 5,120-dimensional hidden representation, processes it with two residual MLP blocks, and predicts a 30-dimensional camera-centric motion vector at each of the 30 action steps.

\paragraph{Action head}
\Cref{fig:action-head-architecture} shows the geometry-conditioned action head. In Stage 2, it is conditioned on ground-truth motion; in Stage 3, it instead receives the motion head's prediction together with pooled VLM hidden states. To efficiently condition the translator on VLM context, we use \textbf{eight learnable query} tokens to attention-pool the full hidden-state sequence from the last layer. The VLM-conditioned features, camera pose, Jacobian, and motion are independently normalized and projected to a shared 2,560-dimensional space. The resulting projected tokens are fused by a two-layer, eight-head Transformer encoder, and its output is mapped to a 30-step, 80-dimensional sparse qpos command chunk.

\subsection{Sparse Command Layout}
\label{app:sparse-command-layout}

The motion and action interfaces use fixed-width tensors so that samples from
different embodiments can share a batch without conflating their physical
semantics. Fixed width does not imply that every channel is valid for every
sample: each target is paired with a dimension-level validity mask, and only
valid entries contribute to the loss or are forwarded to an embodiment's
controller.

\begin{table}[ht]
\centering
\small
\setlength{\tabcolsep}{4pt}
\caption{Canonical 30-dimensional camera-centric motion layout. Slot ranges are inclusive. The final padding slot in each block is structural and is excluded by the validity mask when unused.}
\label{tab:motion30-appendix}
\begin{tabular}{ccc}
\toprule
Block & Slots & Semantic content \\
\midrule
Left manipulator & $0$--$9$ & $p_0$ XYZ, $p_1$ XYZ, $\sin/\cos$ rotation, gripper, padding \\
Right manipulator & $10$--$19$ & $p_0$ XYZ, $p_1$ XYZ, $\sin/\cos$ rotation, gripper, padding \\
Camera motion & $20$--$29$ & camera XYZ translation, 6D continuous rotation, padding \\
\bottomrule
\end{tabular}
\end{table}

\paragraph{30-dimensional camera-centric motion}
At each prediction step, the motion head produces
$m_h=[m_h^{\mathrm{L}},m_h^{\mathrm{R}},m_h^{\mathrm{cam}}]\in\mathbb{R}^{30}$,
organized as three semantically fixed 10-dimensional blocks; the slot-level assignment is summarized in \Cref{tab:motion30-appendix}. For manipulator
$e\in\{\mathrm{L},\mathrm{R}\}$, the block is
\begin{equation}
m_h^e=\left[
\Delta p_{0,h}^e,\Delta p_{1,h}^e,
\sin(\Delta\psi_h^e),\cos(\Delta\psi_h^e),
\gamma_h^e,0
\right]\in\mathbb{R}^{10},
\end{equation}
where the two three-dimensional anchor displacements describe the wrist/end
effector and grasp center in the observing camera frame, $\Delta\psi_h^e$ is
the in-plane rotation, and $\gamma_h^e$ is the gripper state. The final entry
is reserved padding. Camera motion is packed as
\begin{equation}
m_h^{\mathrm{cam}}=\left[
\Delta t_h^{\mathrm{cam}},r_6(\Delta R_h^{\mathrm{cam}}),0
\right]\in\mathbb{R}^{10},
\end{equation}
with three translation channels, the continuous six-dimensional rotation
representation $r_6(\cdot)$, and one reserved entry. A single-arm sample
masks the unused manipulator block; a fixed-camera sample masks the camera
block; and partially annotated data mask only the unavailable channels. Thus,
padding values are structural placeholders rather than supervised zeros.

\begin{table}[ht]
\centering
\small
\setlength{\tabcolsep}{4pt}
\caption{Canonical 80-dimensional action layout. Slot ranges are inclusive;
each row reserves a 10-dimensional semantic block regardless of the active
degrees of freedom of a particular embodiment.}
\label{tab:joint80-appendix}
\begin{tabular}{ccc}
\toprule
Component & Slots & Block content \\
\midrule
Left arm & $0$--$9$ & joint state or command, up to 10 DoF \\
Left end effector & $10$--$19$ & XYZ position (3), 6D continuous rotation (6), padding (1) \\
Left hand/gripper & $20$--$29$ & hand or gripper commands \\
Right arm & $30$--$39$ & joint state or command, up to 10 DoF \\
Right end effector & $40$--$49$ & XYZ position (3), 6D continuous rotation (6), padding (1) \\
Right hand/gripper & $50$--$59$ & hand or gripper commands \\
Waist & $60$--$69$ & waist state or command, interface-dependent DoF \\
Mobile base & $70$--$79$ & base state or command, interface-dependent DoF \\
\bottomrule
\end{tabular}
\end{table}

\paragraph{80-dimensional sparse action}
The action translator predicts $u_h\in\mathbb{R}^{80}$ as eight ordered
10-dimensional blocks,
\begin{equation}
u_h=\left[
 u_h^{\mathrm{L\mbox{-}arm}},u_h^{\mathrm{L\mbox{-}ee}},
 u_h^{\mathrm{L\mbox{-}hand}},u_h^{\mathrm{R\mbox{-}arm}},
 u_h^{\mathrm{R\mbox{-}ee}},u_h^{\mathrm{R\mbox{-}hand}},
 u_h^{\mathrm{waist}},u_h^{\mathrm{base}}
\right].
\end{equation}
\Cref{tab:joint80-appendix} gives the global slot semantics. Arm blocks store
joint commands in a predefined joint order. End-effector blocks store XYZ position channels followed by a 6D continuous rotation~\citep{zhou2019continuity} and one padding channel. Hand blocks store hand or gripper state and commands. Waist and mobile-base blocks store the native state or command variables exposed by the target embodiment, in the interface-defined order within their assigned blocks; for embodiments without a mobile base, slots $70$--$79$ are entirely masked.
When a component has fewer than ten controllable degrees of freedom, its valid
values occupy the leading slots and the remainder are padded and masked. The
same mask prevents inactive blocks from affecting optimization and ensures
that only commands supported by the target embodiment are dispatched at
inference time.

\begin{table}[ht]
\centering
\small
\setlength{\tabcolsep}{4pt}
\caption{Dataset-specific activation of the canonical 80-dimensional action blocks. Listed blocks are active; unused blocks and padded dimensions are masked.}
\label{tab:slot-activation}
\begin{tabular}{lll}
\toprule
Dataset & Active blocks & Canonical slots \\
\midrule
LIBERO & Left arm and gripper & $0$--$9$, $20$--$29$ \\
RoboTwin & Left/right arms and grippers & $0$--$9$, $20$--$39$, $50$--$59$ \\
RoboCasa GR-1 & Left/right arms and hands, waist & $0$--$9$, $20$--$39$, $50$--$69$ \\
\bottomrule
\end{tabular}
\end{table}

\paragraph{Slot Activation}
The canonical layout is shared across datasets, while an embodiment-specific mask selects only the blocks supported by the target control interface. Within each active block, values follow the interface-defined order; dimensions beyond the embodiment's degrees of freedom are padded and excluded from both loss computation and control. Entirely inactive blocks are likewise masked and are never dispatched to the controller. \Cref{tab:slot-activation} summarizes the block-level activation used for the simulated benchmarks.

\clearpage
\subsection{Question-Answer Training Examples}
\label{app:qa-examples}

\begin{figure*}[ht]
\centering
\includegraphics[width=\textwidth]{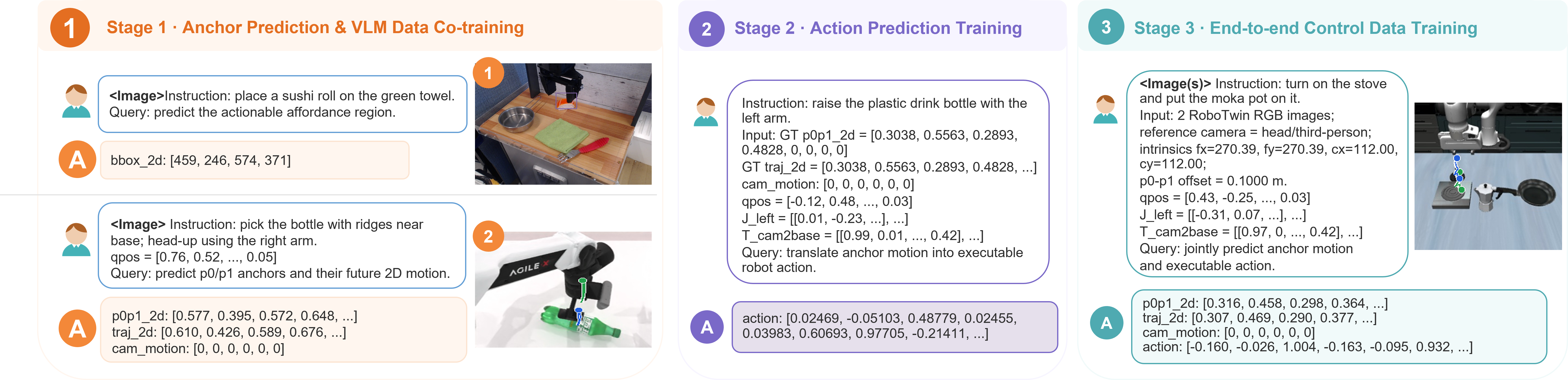}
\caption{
Question-answer examples for the three training stages.
Stage 1 uses visual-language queries for affordance, anchor, and trajectory prediction.
Stage 2 trains action translation from camera-centric motion and robot context.
Stage 3 jointly supervises anchor motion and executable actions with available loss masks.
}
\label{fig:qa-examples}
\end{figure*}

\Cref{fig:qa-examples} illustrates how heterogeneous supervision is expressed through a common question--answer interface. Each sample combines the available observations and task instruction with a query specifying the prediction target. Stage 1 covers affordance grounding, anchor localization, and camera-centric trajectory prediction; Stage 2 maps ground-truth camera-centric motion and robot context to an executable action chunk; and Stage 3 jointly predicts shared anchor motion and embodiment-specific commands. Semantic answers use concise natural language, whereas geometric and control targets use structured numerical outputs, including coordinates for $p_0$ and $p_1$. Source-specific validity masks exclude unavailable targets and inactive command dimensions, allowing visual-language, human, simulation, and real-robot samples to share one interface without treating missing annotations as supervision.

\subsection{Evaluation Protocols}
\label{sec:real-world-protocol}

The following protocols isolate changes in embodiment and data distribution while keeping checkpoint selection, task definitions, and execution conditions controlled.

\paragraph{Simulation}
All simulation benchmarks use the same final checkpoint without benchmark-specific fine-tuning. LIBERO reports the four standard suites and LIBERO-Plus distribution shifts~\citep{liu2024libero}. RoboTwin 2.0 reports Easy and Hard results on ALOHA; the ARX transfer setting fixes scenes, object layouts, perturbation seeds, and camera extrinsics, with initial end-effector poses aligned by inverse kinematics~\citep{mu2024robotwin}. RoboCasa GR-1 evaluates executable humanoid control with arms, hands, and waist commands~\citep{nasiriany2024robocasa}.

\paragraph{Real-world evaluation}
All experiments use Piper robots with 100 demonstrations per task and 20 closed-loop trials. The demonstration budget, optimization schedule, observations, controller, and success criteria are matched across comparisons. Bread pickup tests human-to-robot transfer by converting MediaPipe hand keypoints into camera-centric anchors~\citep{Lugaresi2019MediaPipe}; \Cref{app:human-transfer-details} provides the conversion details. Drawer opening and bowl stacking test SFT adaptation from robot demonstrations, including contact-rich interaction and coordinated bimanual sequencing.

\paragraph{Baselines and metrics}
Simulation baselines use reported results under the closest matching benchmark protocols. Real-world comparisons evaluate $\pi_0$~\citep{black2024pi0} and $\pi_{0.5}$~\citep{physicalintelligence2025pi05} on the same Piper platform under matched deployment conditions. Closed-loop task success is the primary metric. We report average success for LIBERO, RoboTwin, and RoboCasa GR-1, and the mean over seven perturbation categories for LIBERO-Plus. Open-loop prediction errors are used only as supporting diagnostics.

% \FloatBarrier
\clearpage
\section{More Experimental Results}
\label{app:extended-results}

This section reports per-category and per-task results that complement the aggregate comparisons in the main text.

\subsection{Transfer to a New Embodiment}
\label{sec:generalization-transfer}

We isolate embodiment shift by replacing ALOHA with ARX while holding scenes, object layouts, perturbation seeds, and camera extrinsics fixed. Initial end-effector poses are aligned by inverse kinematics. Under this controlled replacement, \method{} achieves 35.0\% success on ARX, as shown in \Cref{tab:arx-transfer}.

\begin{table}[!ht]
\centering
% \small
\setlength{\tabcolsep}{6pt}
\caption{RoboTwin performance under the source ALOHA embodiment and the ARX cross-embodiment setting.}
\label{tab:arx-transfer}
\begin{tabular}{llc}
\toprule
Evaluation setting & Robot & Success (\%) \\
\midrule
Standard Easy & ALOHA & 88.66 \\
Standard Hard & ALOHA & 89.20 \\
Cross-embodiment & ARX & 35.0 \\
\bottomrule
\end{tabular}
\end{table}

\subsection{Ablation on Pooled VLM Features}
\label{sec:pooling-ablation-appendix}

We evaluate whether the action head benefits from visual-language context. The baseline excludes VLM features, whereas the full model conditions the action head on pooled VLM hidden states in addition to motion and geometric inputs. As shown in \Cref{tab:pooling-ablation}, adding the pooled VLM features improves the success rate on RoboCasa GR-1 from 58.3\% to 62.0\%, a gain of 3.7 percentage points. This result shows that the compact VLM context provides complementary visual and language information for downstream action prediction.

\begin{table}[ht]
\centering
\small
\setlength{\tabcolsep}{5pt}
\caption{Effect of adding pooled VLM features to the action head on RoboCasa GR-1.}
\label{tab:pooling-ablation}
\begin{tabular}{lc}
\toprule
Action-head conditioning & RoboCasa GR-1 success rate (\%) \\
\midrule
W/o VLM features         & 58.3 \\
Attend-pooled VLM features & \textbf{62.0} \\
\bottomrule
\end{tabular}
\end{table}

\begin{table*}[h]
\centering
% \scriptsize
\setlength{\tabcolsep}{5pt}
\caption{Success rates (\%) on the four standard LIBERO suites.}
% \vspace{-1em}
\label{tab:libero-suite-breakdown}
\resizebox{0.70\textwidth}{!}
{
\begin{tabular}{lccccc}
\toprule
Method & Spatial & Object & Goal & Long & Average \\
\midrule
Diffusion Policy~{\scriptsize\citep{chi2023diffusionpolicy}} & 78.5 & 87.5 & 73.5 & 64.8 & 76.1 \\
OpenVLA~{\scriptsize\citep{kim2024openvla}} & 84.7 & 88.4 & 79.2 & 53.7 & 76.5 \\
SpatialVLA~{\scriptsize\citep{qu2025spatialvla}} & 88.2 & 89.9 & 78.6 & 55.5 & 78.1 \\
CoT-VLA~{\scriptsize\citep{zhao2025cotvla}} & 87.5 & 91.6 & 87.6 & 69.0 & 83.9 \\
$\pi_0$-FAST~{\scriptsize\citep{black2025fast}} & 96.4 & 96.8 & 88.6 & 60.2 & 85.5 \\
$\pi_0$~{\scriptsize\citep{black2024pi0}} & 96.8 & 98.8 & 95.8 & 85.2 & 94.2 \\
NORA-1.5~{\scriptsize\citep{hung2025nora}} & 97.3 & 96.4 & 94.5 & 89.6 & 94.5 \\
$\pi_{0.5}$~{\scriptsize\citep{physicalintelligence2025pi05}} & 98.8 & 98.2 & 98.0 & 92.4 & 96.9 \\
GR00T-N1.6~{\scriptsize\citep{bjorck2025groot}} & 97.7 & 98.5 & 97.5 & 94.4 & 97.0 \\
OpenVLA-OFT~{\scriptsize\citep{kim2025openvlaoft}} & 97.6 & 98.4 & 97.9 & 94.5 & 97.1 \\
MolmoAct2~{\scriptsize\citep{fang2026molmoact2}} & 97.8 & \textbf{100.0} & 97.8 & 93.2 & 97.2 \\
OASIS~{\scriptsize\citep{chen2026oasis}} & \textbf{99.0} & 98.8 & 97.4 & 95.2 & 97.6 \\
ZR-0~{\scriptsize\citep{li2026zr0}} & 97.4 & 99.4 & 98.0 & 96.4 & 97.8 \\
X-VLA~{\scriptsize\citep{zheng2025xvla}} & 98.2 & 98.6 & 97.8 & \textbf{97.6} & 98.1 \\
MolmoAct2-Think~{\scriptsize\citep{fang2026molmoact2}} & 98.8 & 99.8 & 98.5 & 95.4 & 98.1 \\
ABot-M0~{\scriptsize\citep{yang2026abotm0}} & 98.8 & 99.8 & 99.0 & 96.6 & \textbf{98.6} \\
\midrule
\method{} & 98.8 & 98.6 & \textbf{99.2} & 96.4 & 98.25 \\
\bottomrule
\end{tabular}
}
\end{table*}

\begin{table*}[!t]
\centering
\tiny
\renewcommand{\arraystretch}{0.87}
\setlength{\tabcolsep}{3.5pt}
\caption{Task-level success rates (\%) on RoboCasa GR-1.}
\label{tab:robocasa-gr1-breakdown}
\resizebox{\textwidth}{!}{%
\begin{tabular}{lccccc}
\toprule
Task & \shortstack{GR00T-N1.6\\[-1pt]{\tiny\citep{bjorck2025groot}}} & \shortstack{VP-VLA\\[-1pt]{\tiny\citep{wang2026vpvla}}} & \shortstack{ABot-M0\\[-1pt]{\tiny\citep{yang2026abotm0}}} & \shortstack{JoyAI-RA\\[-1pt]{\tiny\citep{zhang2026joyaira}}} & \method{} \\
\midrule
\texttt{BottleToCabinetClose} & 51.5 & 54.0 & 86.0 & 84.0 & 82.0 \\
\texttt{CanToDrawerClose} & 13.0 & 72.0 & 74.0 & 90.0 & 68.0 \\
\texttt{CupToDrawerClose} & 8.5 & 44.0 & 48.0 & 48.0 & 52.0 \\
\texttt{MilkToMicrowaveClose} & 14.0 & 74.0 & 46.0 & 84.0 & 44.0 \\
\texttt{PotatoToMicrowaveClose} & 41.5 & 34.0 & 50.0 & 70.0 & 60.0 \\
\texttt{WineToCabinetClose} & 16.5 & 48.0 & 66.0 & 54.0 & 62.0 \\
\texttt{CuttingboardToBasket} & 58.0 & 66.0 & 70.0 & 88.0 & 68.0 \\
\texttt{CuttingboardToCardboardbox} & 46.5 & 54.0 & 58.0 & 46.0 & 60.0 \\
\texttt{CuttingboardToPan} & 68.5 & 74.0 & 76.0 & 92.0 & 84.0 \\
\texttt{CuttingboardToPot} & 65.0 & 54.0 & 66.0 & 80.0 & 62.0 \\
\texttt{CuttingboardToTieredbasket} & 46.5 & 56.0 & 38.0 & 36.0 & 38.0 \\
\texttt{PlacematToBasket} & 58.5 & 48.0 & 52.0 & 76.0 & 50.0 \\
\texttt{PlacematToBowl} & 57.5 & 74.0 & 66.0 & 52.0 & 68.0 \\
\texttt{PlacematToPlate} & 63.0 & 70.0 & 60.0 & 38.0 & 60.0 \\
\texttt{PlacematToTieredshelf} & 28.5 & 26.0 & 26.0 & 14.0 & 26.0 \\
\texttt{PlateToBowl} & 57.0 & 52.0 & 54.0 & 48.0 & 58.0 \\
\texttt{PlateToCardboardbox} & 43.5 & 44.0 & 48.0 & 38.0 & 52.0 \\
\texttt{PlateToPan} & 51.0 & 56.0 & 66.0 & 46.0 & 68.0 \\
\texttt{PlateToPlate} & 78.7 & 62.0 & 64.0 & 88.0 & 86.0 \\
\texttt{TrayToCardboardbox} & 51.5 & 44.0 & 54.0 & 82.0 & 72.0 \\
\texttt{TrayToPlate} & 71.0 & 66.0 & 68.0 & 88.0 & 86.0 \\
\texttt{TrayToPot} & 64.5 & 38.0 & 64.0 & 88.0 & 82.0 \\
\texttt{TrayToTieredbasket} & 57.0 & 58.0 & 60.0 & 62.0 & 61.0 \\
\texttt{TrayToTieredshelf} & 31.5 & 24.0 & 38.0 & 24.0 & 40.0 \\
\midrule
Average & 47.6 & 53.8 & 58.3 & 63.2 & 62.0 \\
\bottomrule
\end{tabular}%
}
\end{table*}

\subsection{Full LIBERO Suite Results}
\label{sec:libero-suite-appendix}

\Cref{tab:libero-suite-breakdown} compares success rates on the four standard LIBERO suites. The average is the arithmetic mean of the Spatial, Object, Goal, and Long-Horizon scores. \method{} achieves the best Goal score at 99.2\% and an average of 98.25\%, only 0.35 percentage points below the highest reported average. Its scores remain within a 2.8-point range across all four suites, indicating consistent performance across spatial reasoning, object interaction, goal following, and long-horizon execution. The remaining gap is concentrated in Long-Horizon tasks, where \method{} reaches 96.4\% compared with the best result of 97.6\%.

\subsection{Full RoboCasa GR-1 Task-Level Results}
\label{sec:robocasa-gr1-appendix}

\Cref{tab:robocasa-gr1-breakdown} reports the task-level RoboCasa GR-1 results underlying the aggregate comparison in the main paper. Across the 24 tasks, \method{} obtains an average success rate of 62.0\%, ranking second and trailing JoyAI-RA by 1.2 percentage points. It exceeds ABot-M0, VP-VLA, and GR00T-N1.6 by 3.7, 8.2, and 14.4 points, respectively. \method{} attains the highest task-level success on six tasks, including cup-to-drawer, cuttingboard-to-cardboardbox, and several plate-placement tasks. These results indicate that the shared representation transfers effectively to humanoid control, while the variation across tasks leaves room for stronger contact-rich and articulated-object execution.

\clearpage
\subsection{Full RoboTwin Task-Level Results}
\label{sec:robotwin-appendix}

\Cref{tab:robotwin-breakdown} reports the task-level RoboTwin evaluation results in the same Clean/Randomized layout as prior RoboTwin comparisons. Each task is evaluated over 100 episodes, and we report success rate (SR, \%). \method{} reaches 88.66\% on Clean and the best Randomized average of 89.20\%. The Clean result is within 0.04 points of the strongest baseline, while the Randomized result exceeds ZR-0 by 1.22 points and is 0.54 points higher than \method{}'s own Clean score. This small cross-setting variation indicates strong robustness to randomized scene conditions. However, the low scores on \texttt{OpenMicrowave} (11\%/13\%) identify articulated opening as a clear task-specific limitation despite the strong aggregate performance.

\begin{table*}[!ht]
\centering
\tiny
\renewcommand{\arraystretch}{1.0}
\setlength{\tabcolsep}{1.8pt}
\caption{RoboTwin 2.0 task-level evaluation results. We report success rate under Clean and Randomized settings.}
% \vspace{-1em}
\label{tab:robotwin-breakdown}
\resizebox{\textwidth}{!}{%
\begin{tabular}{lcc cc cc cc cc cc cc}
\toprule
 & \multicolumn{2}{c}{$\pi_0$~{\scriptsize\citep{black2024pi0}}} & \multicolumn{2}{c}{$\pi_{0.5}$~{\scriptsize\citep{physicalintelligence2025pi05}}} & \multicolumn{2}{c}{X-VLA~{\scriptsize\citep{zheng2025xvla}}} & \multicolumn{2}{c}{Motus~{\scriptsize\citep{mu2024robotwin}}} & \multicolumn{2}{c}{LingBot-VLA~{\scriptsize\citep{wu2026lingbotvla}}} & \multicolumn{2}{c}{ZR-0~{\scriptsize\citep{li2026zr0}}} & \multicolumn{2}{c}{\method{}} \\
Task & Clean & Rand. & Clean & Rand. & Clean & Rand. & Clean & Rand. & Clean & Rand. & Clean & Rand. & Clean & Rand. \\
\midrule
\texttt{AdjustBottle} & 99 & 95 & 100 & 99 & 100 & 99 & 89 & 93 & 100 & 100 & 100 & 99 & 100 & 100 \\
\texttt{BeatBlockHammer} & 79 & 84 & 96 & 93 & 92 & 88 & 95 & 88 & 92 & 89 & 85 & 88 & 98 & 96 \\
\texttt{BlocksRankingRGB} & 80 & 63 & 92 & 85 & 83 & 83 & 99 & 97 & 92 & 91 & 92 & 91 & 81 & 91 \\
\texttt{BlocksRankingSize} & 14 & 5 & 49 & 26 & 67 & 74 & 75 & 63 & 76 & 70 & 70 & 81 & 67 & 69 \\
\texttt{ClickAlarmclock} & 77 & 68 & 98 & 89 & 99 & 99 & 100 & 100 & 97 & 43 & 96 & 82 & 78 & 88 \\
\texttt{ClickBell} & 71 & 48 & 99 & 66 & 100 & 100 & 100 & 100 & 43 & 36 & 90 & 83 & 100 & 100 \\
\texttt{DumpBinBigbin} & 88 & 83 & 92 & 97 & 79 & 77 & 95 & 91 & 97 & 97 & 94 & 93 & 89 & 95 \\
\texttt{GrabRoller} & 98 & 94 & 100 & 100 & 100 & 100 & 100 & 100 & 100 & 100 & 100 & 100 & 100 & 100 \\
\texttt{HandoverBlock} & 47 & 31 & 66 & 57 & 73 & 37 & 86 & 73 & 99 & 93 & 93 & 87 & 94 & 92 \\
\texttt{HandoverMic} & 97 & 97 & 98 & 97 & 0 & 0 & 78 & 63 & 100 & 99 & 100 & 99 & 99 & 99 \\
\texttt{HangingMug} & 14 & 11 & 18 & 17 & 23 & 27 & 38 & 38 & 31 & 28 & 35 & 33 & 38 & 39 \\
\texttt{LiftPot} & 80 & 72 & 96 & 85 & 99 & 100 & 96 & 99 & 100 & 99 & 96 & 98 & 100 & 100 \\
\texttt{MoveCanPot} & 68 & 48 & 51 & 55 & 89 & 86 & 34 & 74 & 97 & 87 & 85 & 81 & 92 & 94 \\
\texttt{MovePillbottlePad} & 67 & 46 & 84 & 61 & 73 & 71 & 93 & 96 & 98 & 99 & 98 & 99 & 96 & 91 \\
\texttt{MovePlayingcardAway} & 74 & 65 & 96 & 84 & 93 & 98 & 100 & 96 & 99 & 95 & 99 & 94 & 100 & 99 \\
\texttt{MoveStaplerPad} & 41 & 24 & 56 & 42 & 78 & 73 & 83 & 85 & 93 & 96 & 85 & 92 & 85 & 88 \\
\texttt{OpenLaptop} & 71 & 81 & 90 & 96 & 93 & 100 & 95 & 91 & 96 & 100 & 96 & 99 & 98 & 100 \\
\texttt{OpenMicrowave} & 4 & 32 & 34 & 77 & 79 & 71 & 95 & 91 & 97 & 99 & 94 & 92 & 11 & 13 \\
\texttt{PickDiverseBottles} & 69 & 31 & 81 & 71 & 58 & 36 & 90 & 91 & 85 & 90 & 90 & 88 & 77 & 81 \\
\texttt{PickDualBottles} & 59 & 37 & 93 & 63 & 47 & 36 & 96 & 90 & 95 & 93 & 97 & 98 & 91 & 97 \\
\texttt{PlaceA2BLeft} & 43 & 47 & 87 & 82 & 48 & 49 & 82 & 79 & 99 & 96 & 83 & 80 & 88 & 92 \\
\texttt{PlaceA2BRight} & 39 & 34 & 87 & 84 & 36 & 36 & 90 & 87 & 97 & 92 & 86 & 87 & 91 & 85 \\
\texttt{PlaceBreadBasket} & 62 & 46 & 77 & 64 & 81 & 71 & 91 & 94 & 88 & 91 & 90 & 93 & 94 & 85 \\
\texttt{PlaceBreadSkillet} & 66 & 49 & 85 & 66 & 77 & 67 & 86 & 83 & 92 & 89 & 92 & 85 & 91 & 89 \\
\texttt{PlaceBurgerFries} & 81 & 76 & 94 & 87 & 94 & 94 & 98 & 98 & 99 & 93 & 98 & 98 & 99 & 98 \\
\texttt{PlaceCanBasket} & 55 & 46 & 62 & 62 & 49 & 52 & 81 & 76 & 71 & 73 & 66 & 62 & 78 & 77 \\
\texttt{PlaceCansPlasticbox} & 63 & 45 & 94 & 84 & 97 & 98 & 98 & 94 & 100 & 98 & 86 & 85 & 90 & 94 \\
\texttt{PlaceContainerPlate} & 97 & 92 & 99 & 95 & 97 & 95 & 98 & 99 & 96 & 99 & 99 & 98 & 98 & 100 \\
\texttt{PlaceDualShoes} & 59 & 51 & 75 & 75 & 79 & 88 & 93 & 87 & 90 & 97 & 90 & 95 & 91 & 90 \\
\texttt{PlaceEmptyCup} & 91 & 85 & 100 & 99 & 100 & 98 & 99 & 98 & 100 & 100 & 97 & 97 & 100 & 99 \\
\texttt{PlaceFan} & 66 & 71 & 87 & 85 & 80 & 75 & 91 & 87 & 91 & 92 & 85 & 78 & 94 & 96 \\
\texttt{PlaceMousePad} & 20 & 20 & 60 & 39 & 70 & 70 & 66 & 68 & 89 & 82 & 85 & 83 & 85 & 84 \\
\texttt{PlaceObjectBasket} & 67 & 70 & 80 & 76 & 44 & 39 & 81 & 87 & 90 & 88 & 75 & 77 & 83 & 88 \\
\texttt{PlaceObjectScale} & 57 & 52 & 86 & 80 & 52 & 74 & 88 & 85 & 90 & 87 & 88 & 89 & 89 & 97 \\
\texttt{PlaceObjectStand} & 82 & 68 & 91 & 85 & 86 & 88 & 98 & 97 & 95 & 93 & 90 & 91 & 99 & 92 \\
\texttt{PlacePhoneStand} & 49 & 53 & 81 & 81 & 88 & 87 & 87 & 86 & 95 & 95 & 85 & 81 & 87 & 88 \\
\texttt{PlaceShoe} & 76 & 76 & 92 & 93 & 96 & 95 & 99 & 97 & 99 & 100 & 98 & 97 & 99 & 100 \\
\texttt{PressStapler} & 44 & 37 & 87 & 83 & 92 & 98 & 93 & 98 & 87 & 81 & 90 & 92 & 97 & 99 \\
\texttt{PutBottlesDustbin} & 65 & 56 & 84 & 79 & 74 & 77 & 81 & 79 & 95 & 97 & 82 & 79 & 90 & 87 \\
\texttt{PutObjectCabinet} & 73 & 60 & 80 & 79 & 46 & 48 & 88 & 71 & 87 & 86 & 82 & 76 & 95 & 92 \\
\texttt{RotateQRcode} & 74 & 70 & 89 & 87 & 34 & 33 & 89 & 73 & 83 & 82 & 78 & 85 & 91 & 94 \\
\texttt{ScanObject} & 55 & 42 & 72 & 65 & 14 & 36 & 67 & 66 & 98 & 96 & 86 & 85 & 88 & 98 \\
\texttt{ShakeBottleHorizontally} & 98 & 92 & 99 & 99 & 100 & 100 & 100 & 98 & 100 & 100 & 100 & 100 & 99 & 99 \\
\texttt{ShakeBottle} & 94 & 91 & 99 & 97 & 99 & 100 & 100 & 97 & 100 & 100 & 100 & 100 & 100 & 99 \\
\texttt{StackBlocksThree} & 72 & 52 & 91 & 76 & 6 & 10 & 91 & 95 & 60 & 62 & 86 & 88 & 94 & 92 \\
\texttt{StackBlocksTwo} & 93 & 79 & 97 & 100 & 92 & 87 & 100 & 98 & 95 & 93 & 95 & 91 & 96 & 89 \\
\texttt{StackBowlsThree} & 77 & 75 & 77 & 71 & 76 & 86 & 79 & 87 & 80 & 81 & 79 & 88 & 97 & 95 \\
\texttt{StackBowlsTwo} & 94 & 95 & 95 & 96 & 96 & 93 & 98 & 98 & 95 & 93 & 94 & 92 & 88 & 90 \\
\texttt{StampSeal} & 46 & 33 & 79 & 55 & 76 & 82 & 93 & 92 & 90 & 90 & 92 & 92 & 94 & 87 \\
\texttt{TurnSwitch} & 41 & 42 & 62 & 54 & 40 & 61 & 84 & 78 & 71 & 76 & 83 & 78 & 54 & 53 \\
\midrule
{\bfseries Average} & 65.92 & 58.40 & 82.74 & 76.76 & 72.80 & 72.84 & 88.66 & 87.02 & 88.56 & 86.68 & 88.70 & 87.98 & {\bfseries 88.66} & {\bfseries 89.20} \\
\bottomrule
\end{tabular}
}
\end{table*}
\renewcommand{\arraystretch}{1.0}

% \FloatBarrier

\clearpage
\section{More Visualizations}
\label{appendix-vis}

This section provides qualitative manipulation examples and visualizes the conversion of human-hand observations into the shared camera-centric anchor representation.

% \subsection{LIBERO-Plus Robustness Visualization}

% \Cref{fig:libero-plus-results} visualizes the success rates under the seven LIBERO-Plus perturbation categories reported in \Cref{sec:libero-plus-appendix}.

% \begin{figure*}[t]
% \centering
% \includegraphics[width=\textwidth]{figs/libero_plus_results.pdf}
% \caption{Out-of-distribution robustness across all seven LIBERO-Plus perturbation categories. Total summarizes Camera, Robot, Language, Light, Background, Noise, and Layout success rates.}
% \label{fig:libero-plus-results}
% \end{figure*}

\subsection{Human-to-Robot Conversion Details}
\label{app:human-transfer-details}

\begin{figure}[h]
\centering
\includegraphics[width=\linewidth]{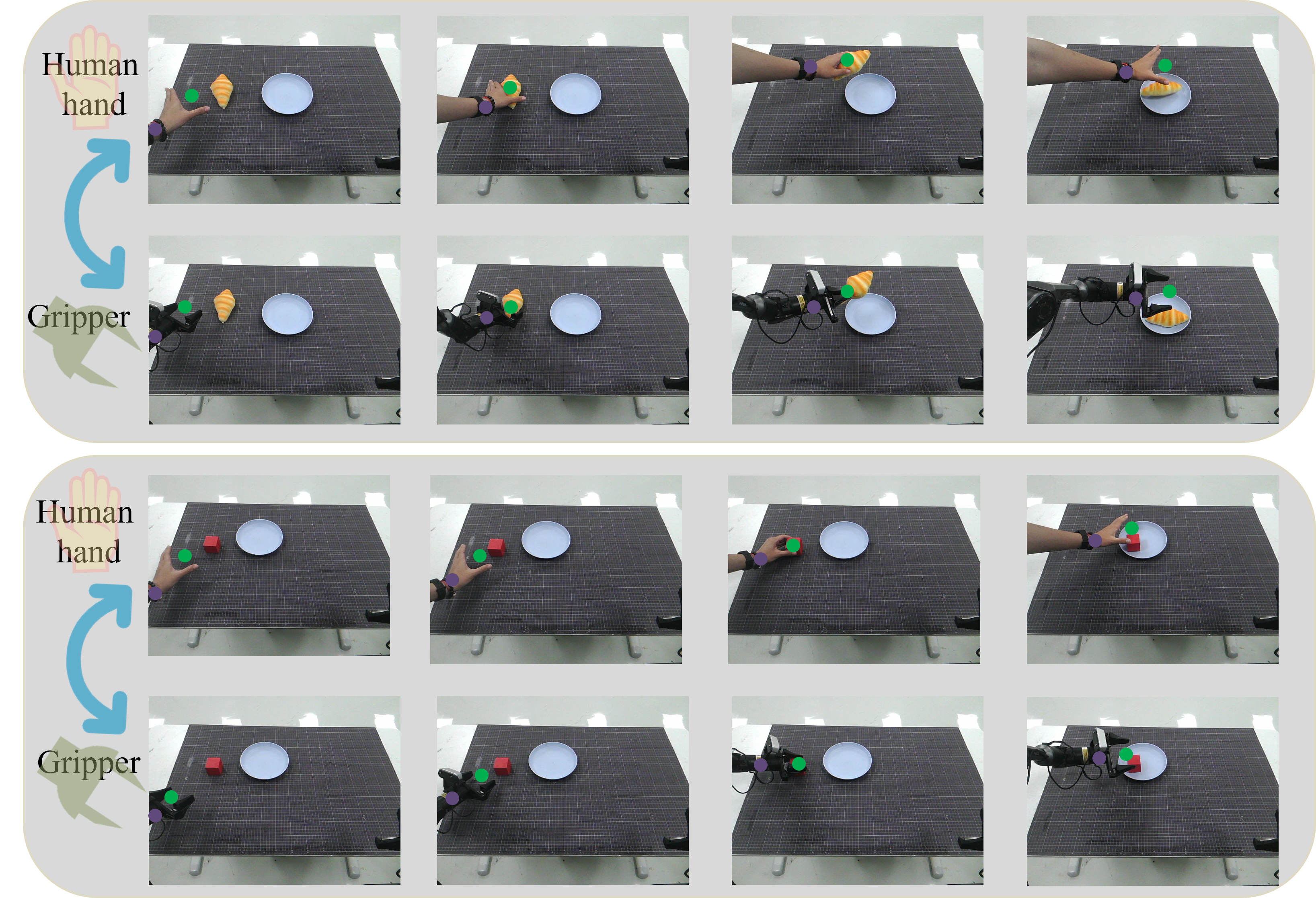}
\caption{
Mapping human-hand demonstrations to robot supervision. Human videos provide camera-centric anchor motion used to train the robot policy.
}
\label{fig:mapping}
\end{figure}

For bread pickup, human-hand demonstrations are converted into the shared camera-centric representation using MediaPipe keypoint detection~\citep{Lugaresi2019MediaPipe}. As shown in \Cref{fig:mapping}, the wrist keypoint is mapped to $p_0$, and the midpoint between the thumb tip and index fingertip is mapped to $p_1$. The resulting anchor trajectories are aligned to the camera frame and used as motion supervision, while unavailable channels remain masked. This conversion preserves the task-relevant approach and grasp geometry visible in the demonstration without assuming a robot-specific kinematic structure. During training, these samples supervise only the shared motion head; executable robot-action slots remain invalid and therefore do not contribute to the action loss.

\clearpage
\subsection{Human Demonstration Example}
\label{app:human_demonstration}

\begin{figure}[!ht]
\centering
\includegraphics[width=\linewidth]{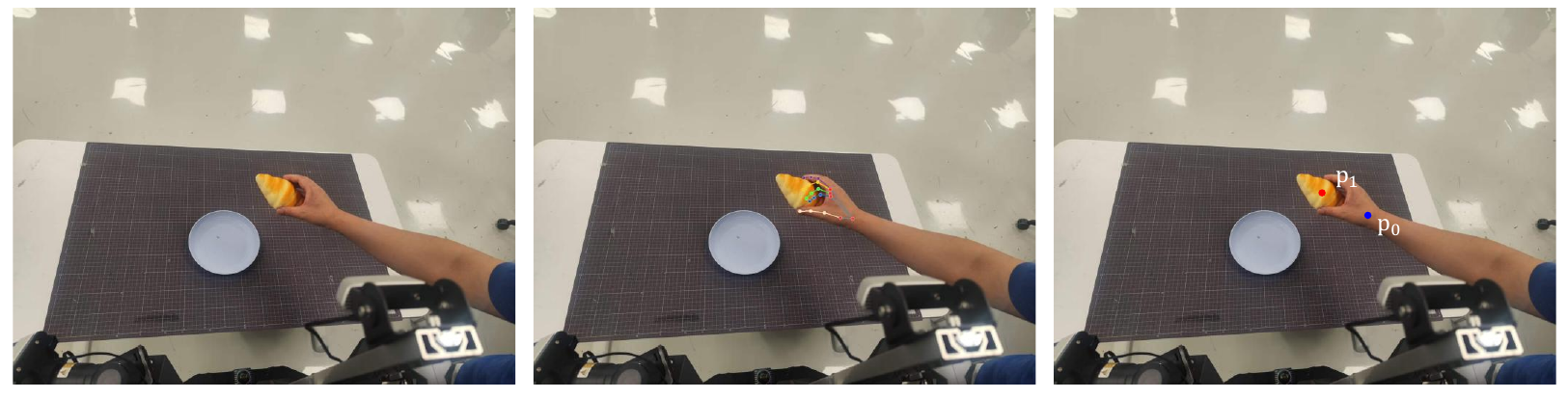}
\caption{
Human-to-robot conversion for the bread-pickup task using MediaPipe~\citep{Lugaresi2019MediaPipe} hand keypoints.
}
\label{fig:hand-transform-appendix}
\end{figure}

\Cref{fig:hand-transform-appendix} illustrates the human-demonstration pipeline, including video collection, hand-keypoint detection, and estimation of the camera-centric anchors $p_0$ and $p_1$ used for robot learning. The example highlights that human videos provide geometric motion supervision without requiring human actions to be expressed in robot joint coordinates.

\subsection{Qualitative Manipulation Examples}
\label{app:qualitative-examples}

\begin{figure*}[!ht]
\centering
\makebox[\textwidth][c]{%
\begin{minipage}[t]{0.55\textwidth}
\centering
\includegraphics[page=1,width=\linewidth]{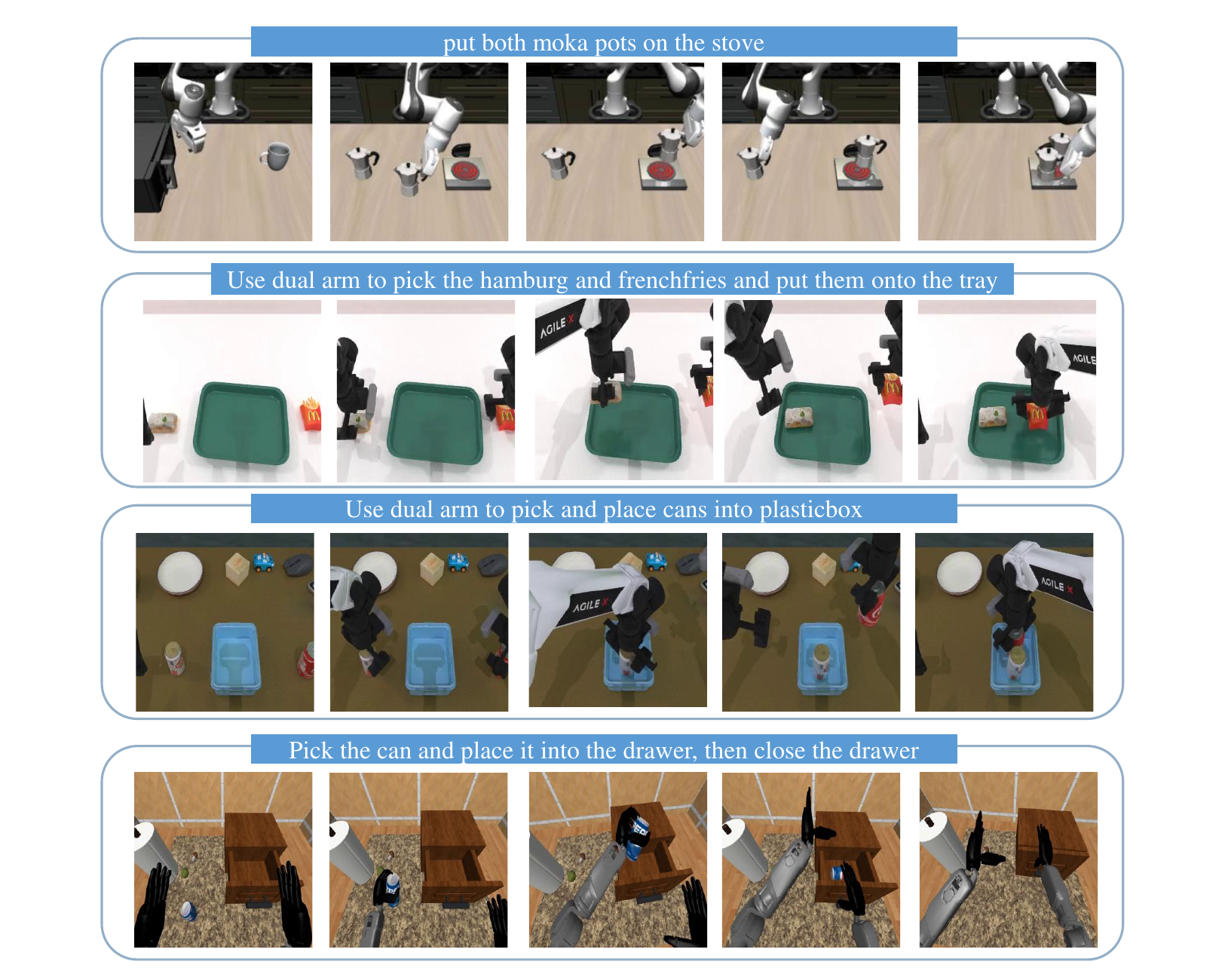}
\small (a) Simulation
\end{minipage}%
\hspace{-0.03\textwidth}%
\begin{minipage}[t]{0.55\textwidth}
\centering
\includegraphics[page=2,width=\linewidth]{figs/examples.pdf}
\small (b) Real-world
\end{minipage}
}
\caption{
Qualitative manipulation examples from (a) simulated benchmarks, including LIBERO, RoboTwin, and RoboCasa GR-1, and (b) real-world Piper robot tasks.
}
\label{fig:qualitative-examples}
\end{figure*}

\Cref{fig:qualitative-examples} shows representative qualitative examples across simulated and real-world manipulation tasks. The paired views illustrate how the same camera-centric representation is reused across controlled simulation and real-world execution, where sensing and contact introduce additional variation.

\newpage
\section{Failure cases and Limitation Analysis}
\label{appendix:failure-analysis}
% We presented \method{}, a camera-centric formulation for learning vision-language-action policies from heterogeneous embodied data.
% By representing manipulation through camera-observable anchor motion and translating the shared geometric prediction into embodiment-specific commands, \method{} decouples transferable task geometry from robot-specific control interfaces.
% Across single-arm, dual-arm, humanoid, out-of-distribution, and real-world evaluations, the same formulation supports unified manipulation, cross-embodiment transfer, human-to-robot transfer, and data-efficient adaptation.
% These results suggest that camera-centric action geometry provides a practical interface for scaling reusable robot policies across heterogeneous embodiments and demonstration sources.

\begin{figure*}[!ht]
\centering
\includegraphics[width=0.8\textwidth]{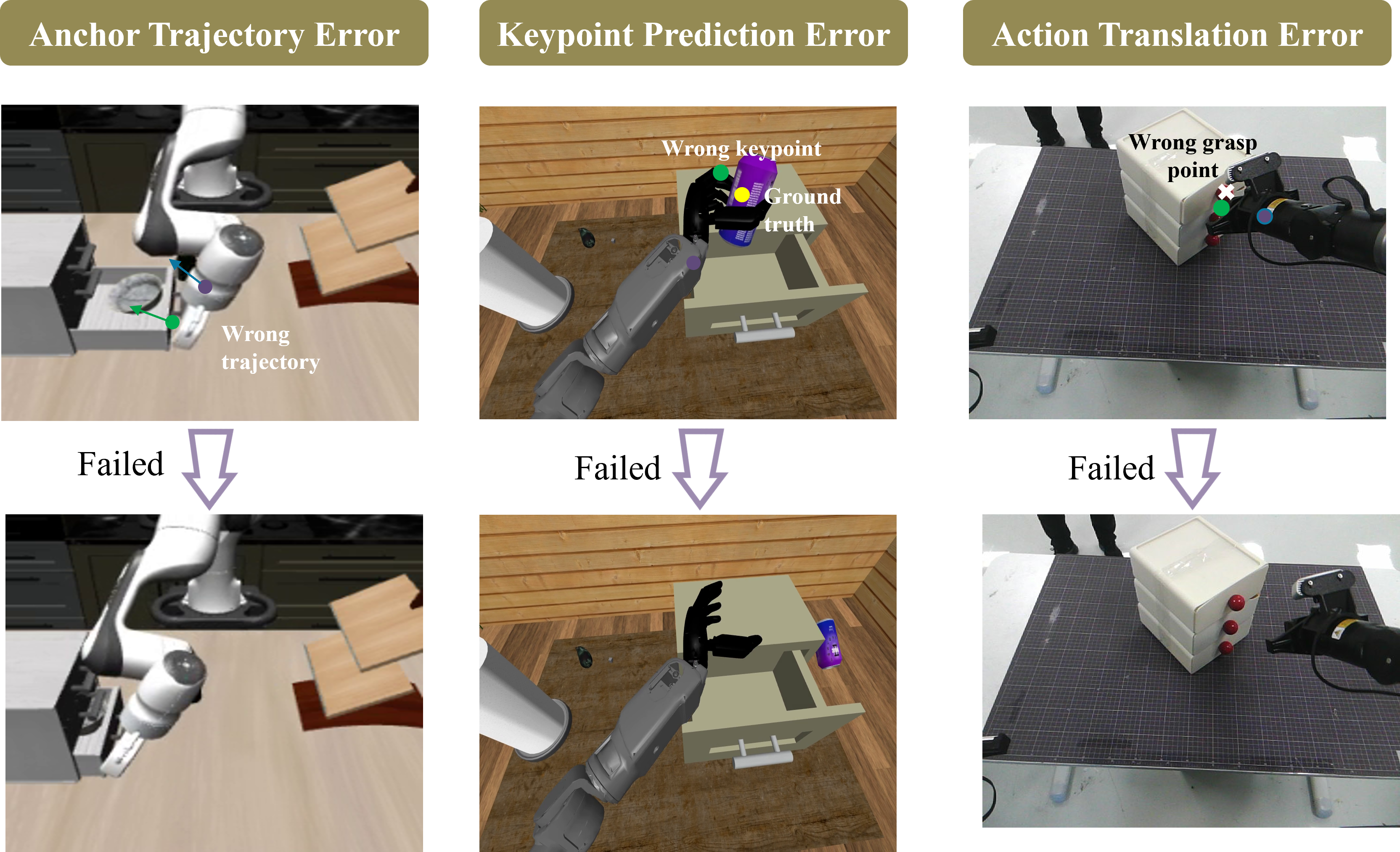}
\caption{
Representative failure cases caused by errors in anchor detection, trajectory prediction, and action translation.}
\label{fig:failed}
\end{figure*}

\Cref{fig:failed} shows representative failure cases arising from errors in anchor detection, trajectory prediction, and action translation.
\method{} has three main limitations. First, it relies on geometric information that must be estimated or calibrated reliably: errors in camera calibration, depth estimation, embodiment kinematics, or MediaPipe hand-keypoint localization can propagate into the camera-centric target and the downstream translator. Second, cross-embodiment transfer remains challenging---although \method{} achieves non-zero ALOHA-to-ARX transfer, the gap from the source embodiment indicates that camera-centric motion alone does not eliminate morphology and kinematic mismatch. Third, our real-world evaluation is intentionally controlled and limited in scale: it supports human-to-robot transfer and data-efficient SFT adaptation, but broader robustness requires evaluation across more robots, camera setups, object categories, and long-horizon tasks.

\end{document}